\documentclass[11pt]{scrartcl}       

\usepackage[bottom=2.5cm,top=2.5cm,left=2.5cm,right=2.5cm]{geometry}

\usepackage{amsmath,amssymb,amsthm,graphicx,caption,cases}	
\usepackage[bottom]{footmisc}	
\usepackage{algorithmic}

\usepackage{abstract,amsfonts,amsbsy,amssymb,amstext,mathrsfs,latexsym,color,url,amsthm,xcolor,authblk,fontenc,bm}

\definecolor{NUSBlue}{RGB}{0,61,124} 
\definecolor{NUSOrange}{RGB}{239,124,0}
\usepackage[colorlinks=false,allbordercolors=NUSOrange]{hyperref}

\DeclareOldFontCommand{\bf}{\normalfont\bfseries}{\mathbf}

\newcommand{\bz}{\mathbf{z}}

\newcommand{\bp}{\mathbf{p}}
\newcommand{\ba}{\mathbf{a}}

\newcommand{\bc}{\mathbf{c}}

\newcommand{\be}{\mathbf{e}}

\newcommand{\bx}{\mathbf{x}}
\newcommand{\by}{\mathbf{y}}

\newcommand{\R}{\mathbb R}

\definecolor{NUSBlue}{RGB}{0,61,124}   

\usepackage{tikz}
\usepackage{mathtools}

\def\nudge{.5}

\tikzset{axis/.style={ultra thin, Grey, -latex, shorten <=-\nudge cm, shorten >=-2*\nudge cm}}
\tikzset{line/.style={thick}}

\usepackage{textcomp}
\DeclareMathAlphabet\mathbfcal{OMS}{cmsy}{b}{n}
\usepackage{hyperref}
\usepackage[linesnumbered,vlined,ruled]{algorithm2e}
\usepackage{amsmath,amsfonts,amssymb,amsthm}
\theoremstyle{plain}
\newtheorem{thm}{Theorem}

\newtheoremstyle{cited}%
  {3pt}
  {3pt}
{\itshape}
  {}
  {\bfseries}
  {.}
  {.5em}
  {\thmname{#1} \thmnumber{#2} \thmnote{\normalfont#3}}
\theoremstyle{cited}
\newtheorem{citedthm}[thm]{Theorem}
\newtheorem{citedlem}[thm]{Lemma}
\newtheorem{citedcor}[thm]{Corollary}

\newtheorem{citedprop}[thm]{Proposition}

\usepackage{graphicx}
\usepackage{epstopdf}
\usepackage{epsf}
\usepackage{subfigure}
\usepackage{epsfig}

\usepackage{cite}

\usepackage{tabularx}
\usepackage{array}
\usepackage{booktabs}

\usepackage{comment}
\usepackage{tgpagella} 
\usepackage{mathpazo}

\begin{document}
\renewcommand*{\Authsep}{, }
\renewcommand*{\Authand}{, }
\renewcommand*{\Authands}{, }
\renewcommand*{\Affilfont}{\normalsize}   
\setlength{\affilsep}{2em}   
\title{Generalization bounds for regression and classification on adaptive covering input domains}
\date{\today}
\author{Wen-Liang Hwang}
\maketitle

\begin{abstract}
The generalization error is the expected test error that compares the outcomes of an unknown target function with a hypothesis function derived from training data. Our primary focus is on the generalization bound, which acts as an upper bound for the generalization error. Our analysis delves into regression and classification tasks separately to ensure a comprehensive examination.
We assume the target function is real-valued and Lipschitz continuous for regression tasks. We measure the differences between predictions and actual values using the $l_2$-norm, a root-mean-square-error (RMSE) variant. As for classification tasks, we treat the target function as a one-hot classifier, representing a piece-wise constant function, and use 0/1 loss for error measurement. We denote $d$ as the dimension of the input domain, the unit $d$-ball. Our analysis shows that to achieve a concentration inequality of generalization bounds denoted by $(\epsilon, \delta)$, where $\epsilon$ signifies the accuracy parameter of the bound and $1-\delta$ represents the probability guarantee of achieving the accuracy, the sample complexity for learning a regression function is ${\cal O}(\frac{1}{\delta \epsilon^d})$, and for the classification function is ${\cal O}(\frac{1}{\delta \epsilon^{\frac{d}{d-1}}})$. This finding highlights the varying learning efficiency for regression and classification tasks.
In addition, we show that the generalization bounds for regression functions are proportional to $\frac{(K_f + K_{\mathcal{M}})}{ \#W_{\mathcal{M}}^{1/\log_2 (1+\alpha)}}$, where $K_f$ and $K_{\mathcal M}$ are the Lipschitz constants of the target and network regression functions, $\alpha \geq 1$ depends on the hypothesis class and the network architecture, and $\#W_{\mathcal M}$ represents the number of network parameters. On the other hand, the generalization bounds for classification functions are proportional to $\frac{ |\partial f|}{\#W_{\mathcal M}^{(d-1)/\log_2(1 + \alpha)}}$. Here, $|\partial f|$ denotes the length of the classification boundary of the classification function $f$. These results highlight the advantages of over-parameterized networks, as the generalization bound is inversely proportional to the number of parameters of a network.
\end{abstract}


\section{Introduction}

Dealing with the complexities of function families in machine learning involves addressing infinite hypothesis sets by converting them into finite ones. This process results in bounding generalization errors for infinite hypothesis spaces. The generalization error is the expected test error of matching the outputs between an unknown target function and a hypothesis function derived from training data. The primary focus of studying is on the generalization bound, which provides an upper bound for the generalization error for all hypothesis functions in the space of interest. 
The technique of covering number focuses on determining the minimum number of balls required to cover a family of hypothesis functions, while the packing number deals with the maximum number of non-overlapping balls in the hypothesis space.
Growth functions, Radamacher's complexity, and VC dimensions are integral to this understanding. We refer to sources such as \cite{mohri2018foundations,anthony1999neural} and other notable books and lecture notes to further explore the methods and their relationships. 

The generalization bounds obtained from analyzing the VC dimension of neural networks for classification include a specific term proportional to the network size. The analysis indicates that as network size increases, the generalization bound also increases, resulting in a trivial bound for generalization error when the number of parameters exceeds the training data \cite{baum1988size,bartlett2019nearly,bartlett1998almost}. These findings have underscored a series of research efforts in the departure of VC dimension by exploring generalization bounds independent of network size to minimize the impact of the number of layers and weights using the magnitude of network parameters  \cite{bartlett2017spectrally,Golowich2018size}, PAC-Bayesian \cite{neyshabur2015norm,neyshabur2017pac}, implicit regularization, and norm-based inductive bias.

The above approach does not take advantage of the analytical results, indicating that a neural network partitions its input domain into smaller segments. Each segment is tailored to the local geometry of the target function, conditioned on the training data and the activation functions of the network. This concept is supported by Pascanu et al. \cite{pascanu2013number}, Hwang et al. \cite{hwang2023analysis}, and Kadkhodaie et al. \cite{kadkhodaie2023generalization}, as well as by the "double-descent phenomenon" identified by Belkin et al. \cite{belkin2019reconciling}, which indicates a clear correlation between the number of weights and generalization errors. Moreover, Bartlett et al. \cite{bartlett2020benign} found similar relationships in ridge regression analysis, emphasizing the significance of over-parameterization in minimizing generalization errors. 
These discoveries are also reinforced within the context of constructing a deep neural network for the universal approximation theory \cite{heinecke2020refinement}. The theory demonstrates a sparse convolution network, utilizing parameters introduced by adding more layers or levels to the neural network to more effectively approximate a function by offering a finer local structure within the input domain.

Our analysis is conducted to align with the previously mentioned findings. We aim to integrate local geometric segments a network induces into our analysis to establish a generalization bound that diminishes as the network's size increases. We conduct thorough and separate analyses for regression and classification tasks to ensure a comprehensive examination. We assume the target function is real-valued and Lipschitz continuous for regression tasks. We measure the deviations between predictions and true values using the $l_2$-norm, a root-mean-square-error (RMSE) variant. For classification tasks, we treat the target function as a one-hot classifier, representing a piece-wise constant function, and use 0/1 loss for error measurement.

\subsection{Contributions}

Our efforts lead to the following contributions:

$\bullet$ 
Our study uses the geometric parameters $\gamma_s$ to characterize the generalization errors for regression and classification tasks. The parameter $\gamma_s$ represents the smallest radius of a ball that can flexibly encompass the shape of local geometric regions within a network's input domain. We also analyze the significance of the value of $\gamma_s$ concerning generalization bounds for each task. For learning regression function $f$ with Lipschitz constant $K_f$ using machine $\mathcal M$ with Lipschitz constant $K_{\mathcal M}$, the bound is $(K_f + K_{\mathcal M}) \gamma_s$. Additionally, for one-hot classification function $f$ with classification boundary $\partial f$, the bound is $(\gamma_s)^{d-1} \text{vol } B_{d-1}(\b0, 1) | \partial f|$, where $d$ is the dimension of the input domain, $|\partial f|$ denotes the length of $\partial f$, and $ \text{ vol} B_{d-1}(\b0, 1) $ denotes the volume of unit $(d-1)$-ball. \\

$\bullet$ 
Our analysis established a relationship between the parameter value $\gamma_s$ and the number of network parameters. Specifically, we found that $\gamma_s$ is inversely proportional to a polynomial of the parameter count of a network. The polynomial's order varies depending on the algorithm and network architecture. This outcome aligns with the understanding that increasing the network's layers leads to a larger parameter count, resulting in a smaller value of $\gamma_s$ due to the refinement of the network's input domain. Building on this analysis, we have developed a deep neural network with $\gamma_s \leq \frac{1}{\#W_L}$, where $\#W_L$ represents the parameters of a network of layer $L$. \\

$\bullet$ 
The results mentioned above are determined based on two assumptions about the density of the training data in the input domain sampled from an unknown distribution. We express these assumptions statistically to demonstrate using concentration inequalities with precision parameter $\epsilon$ and confidence parameter $\delta$ of the generalization bounds. We then show that to achieve the desired concentration bound $(\epsilon, \delta)$ with $\delta > \exp^{-\frac{\epsilon^2}{2}}$, which is the upper bound of generalization error with precision $\epsilon$ with confidence larger than $1-\delta$, the sample complexity for the regression function is ${\cal O}(\frac{1}{\delta \epsilon^d})$. In contrast, for the classification function, it is ${\cal O}(\frac{1}{\delta \epsilon^{\frac{d}{d-1}}})$. \\

$\bullet$ 
In light of our findings, we address two unresolved theoretical issues: benign overfitting in interpolating networks and the varying effectiveness of learning on regression and classification tasks. \\

The remainder of this document is organized as follows: We begin by reviewing previous research on bounding generalization errors in Section \ref{sec:review}. Next, in Section \ref{sec:main}, we provide detailed information about the main findings of our study. Following that, in Section \ref{oracle}, we introduce an implementation of a deep neural network that links the number of parameters to the local geometric structure of the input domain of the network. In Section \ref{sec:impact}, we explain experimental findings about different learning efficiencies for regression and classification tasks and benign overfitting.
Finally, in Section \ref{sec:discussion}, we discuss technical comments and provide suggestions to enhance our approach.


\section{Related works} \label{sec:review}

Understanding generalization bounds is crucial for gaining insight into the fundamental properties of deep neural networks. Numerous researchers have focused on studying this complex issue, and it's important to stay updated on the latest progress in the literature. In this review, we will outline some of the research contributions in this area that we are aware of.

The VC dimension analysis for a neural network used for classification with $L$ layers and $\#W$ weights varies depending on the types of activation functions. For linear threshold and piece-wise linear (ReLU-type) units, the VC dimension is approximately $\#W$ \cite{baum1988size} and $\#WL$ \cite{bartlett2019nearly}, respectively. In contrast, piece-wise polynomial units have a VC dimension of approximately $\#WL^2$ \cite{bartlett1998almost}.

The VC dimension for a neural network with $L$ layers and $\#W$ weights varies based on the types of activation functions. For linear threshold and piece-wise linear (ReLU-type) units, the VC dimension is approximately $\#W$ \cite{baum1988size} and $\#WL$ \cite{bartlett2019nearly}, respectively. In contrast, piece-wise polynomial units have a VC dimension of approximately $\#WL^2$ \cite{bartlett1998almost}.

Recent research suggests that incorporating scale-sensitive bounds, which consider the margin concept, Rademacher complexity, and weight coefficient magnitude, can help reduce the reliance of generalization bounds on the width and depth of neural networks  \cite{bartlett2017spectrally}. This approach emphasizes using norm-based constraints to control the complexity of the hypothesis space. Initially, Neyshabur et al. \cite{neyshabur2015norm} proposed an exponential bound for depth $L$, which were later improved by Bartlett et al. \cite{bartlett2017spectrally} to polynomial bounds using spectrally-normalized margin bounds. Neyshabur et al. \cite{neyshabur2017pac} extended this to PAC-Bayesian bounds. At the same time, Golowich et al. \cite{Golowich2018size} made the bounds independent of $L$ by assuming appropriate norm constraints and depth-independent sample complexity for the network class. Golowich et al.'s improvements are significant as the dependence of generalization bounds to depth suggests that shallow neural networks possess better generalization capabilities than deep networks, contrary to common belief. Moreover, the work also comprehensively compares various bounds for scale-sensitive and PAC-Bayesian approaches.

When dealing with over-parameterized models, an explicit inductive bias is frequently applied to the objective function to obtain a regularized solution. For instance, the research in \cite{yang2024nearly} explores loss functions of neural networks that incorporate function derivatives and provides a comprehensive VC-dimension-based generalization bound in the Sobolev space. Additionally, optimization techniques may inherently lean towards favoring a regularized solution. A notable example is the convergence to the $l_2$-regularized solution of employing the gradient method on over-parameterized supervised learning linear systems when the initial weight of the gradient method is set to the zero vector \cite{zhang2021understanding}. The smaller generalization error of a parametric model can also be achieved through an algorithmic approach, as long as the training algorithm is stable, according to Bousquet and Elisseeff \cite{bousquet2002stability}. In simpler terms, this implies that the training error generated by the algorithm changes only slightly when a single data point in the training set is modified. This property holds for smooth, and Lipschitz functions, making it particularly relevant for stochastic gradient descent \cite{hardt2016train}. 
Recent research has investigated the implicit regularization of stochastic gradient descent and its impact on generalization errors in various learning objectives for neural networks. These studies have explored different neural network architectures, such as wide neural networks known for their unique characteristics \cite{lee2019wide}, as well as single-hidden layer neural networks using square loss with various activation functions in the hidden layer, including quadratic \cite{li2018algorithmic}, monotonic, and ReLU activation functions \cite{vardi2021implicit}. Additionally, the work by \cite{wang2022generalization} illustrates algorithm-specific generalization error bounds by combining the trajectory of stochastic gradient descent with norm-based generalization error estimates, assuming that the network function is bounded.

\section{Main results} \label{sec:main}

The analysis in \cite{neyshabur2015norm,bartlett2017spectrally} assumes the input domain  is $\|\bx\|_2 \leq B$ for some $B$. The following analysis suggests using balls to cover the unit ball in $\mathbb{R}^d$. This suggestion ensures a finite number of coverings, and we will show that it can maintain generalization within certain limits. Using a unit $d$-ball for the input domain does not limit our analysis, as it can be easily adjusted to include any compact input domain commonly encountered in real-world applications.

The behavior of a neural network's function adapts to its geometry within the input domain \cite{pascanu2013number, hwang2023analysis,kadkhodaie2023generalization}. We can understand the local geometry of a function by using the concept of smallest enclosing balls, where the radius of these balls represents the generalization bound. By making specific assumptions about the distribution of training points within local regions, we can determine the generalization bound without knowing the input density function (see Lemmas \ref{generalizationbound} and \ref{generalizationboundc}). Furthermore, we establish that the generalization bound is inversely proportional to a polynomial of the number of parameters in the network by examining the relationship between the number of parameters and the radius of the enclosing balls (see Propositions \ref{prop1} and \ref{prop2}).

Next, we express each assumption as a statistical statement with accuracy and confidence parameters as a concentration inequality. Combining these concentration inequalities allows us to derive the generalization bounds for regression and classification functions (see Theorems \ref{generalizationboundRB} and \ref{generalizationboundCB}) and their corresponding sample complexities (see Corollaries \ref{samplecomplexityR} and \ref{samplecomplexityC}).

\subsection{Generalization bound for covering with fixed radius deterministic balls}

We make two assumptions on training data in our generalization error analysis. Firstly, we assume that the data is distributed densely enough to cover the input domain within a certain radius, which we refer to as the "assumption of densely covering." To access the density of training data in the input domain, we use the $\gamma$-covering of the input domain, which means that for any $\bx$ in the input domain, there is a training point $\bx_i$ such as $\|\bx - \bx_i\|_2 \leq \gamma$. The concept of $\gamma$-dense distribution in training points emphasizes that the input domain can be covered using balls of radius $\gamma$, and each ball contains at least one training point.

Secondly, we assume that enough training data is available for each covered area to allow for an accurate approximation of probability mass using a histogram distribution. This is referred to as the "assumption of histogram approximation." These two assumptions form the basis of our analysis and help us identify the key parameters contributing to generalization errors concerning domain covering.

\subsubsection{Regression networks with Lipschitz continuity}

The target function $f: B_d(\b0, 1) \rightarrow \R^l$ is a Lipschitz function wit Lipschitz constant $K_f$ where $B_d(\b0, 1)$ denotes the unit $d$-ball (i.e., the unit ball centered about $\b0$ in $\R^d$). The learning algorithm works on input distribution $(\bx, f(\bx))$, defined as $B_d(\b0,1) \times \R^l$-valued regression pairs, where $\bx$ has a density $\mathcal P$. 
Suppose network $\mathcal M$ is a continuous function of Lipschitz constant $K_{\mathcal M}$, and the network is derived from $\gamma_s$-dense distributed training data $\{(\bx_i, f(\bx_i)\}_i$ by minimizing the empirical loss function (that is, the expected error of the training data).

For any input point $\bx \in B_d(\b0,1)$, the error between $f(\bx)$ and $\mathcal M(\bx)$ can bounded by the distance between the input point and training input $\bx_i$ as  
\begin{align} \label{maingen}
\|f(\bx) - \mathcal M(\bx) \|_2 & =  \|f(\bx) - \mathcal M(\bx) + f(\bx_i) - f(\bx_i) + \mathcal M(\bx_i) - \mathcal M(\bx_i)\|_2 \nonumber \\
& \leq \| f(\bx) - f(\bx_i)\|_2 + \| f(\bx_i) - \mathcal M(\bx_i) \|_2+\|  \mathcal M(\bx_i) - \mathcal M(\bx)\|_2 \nonumber \\
& \leq K_f \| \bx - \bx_i \|_2 + \varepsilon_i + K_{\mathcal M} \| \bx - \bx_i \|_2.
\end{align}
The first and third terms in the right-hand side come from the Lipschitz continuity of $f$ and $\mathcal M$, respectively, and $\varepsilon_i$ is the 2-norm error between $f$ and $\mathcal M$ at $\bx_i$.
Because the training data set is $\gamma_s$-dense in $B_d(\b0,1)$,  for any $\bx \in B_d(\b0,1)$, there is a training input $\bx_p$ such that $\| \bx - \bx_p \|_2 \leq \gamma_s$. Now, substituting $\bx_p$ for $\bx_i$ in Eq. (\ref{maingen}) and using $\| \bx - \bx_p \|_2 \leq \gamma_s$, we obtain
\begin{align} \label{uniformbound}
\|f(\bx) - \mathcal M(\bx) \|_2  \leq  (K_f + K_{\mathcal M}) \gamma_s + \varepsilon_p.
\end{align}

If the set of training points $\{\bx_i\}$ forms a $\gamma_s$-covering of the input domain, it contains a subset, denoted as $C$. For each $\bc_i \in C$, where $p_i = B_d(\bc_i, \gamma_s)$ denotes a ball of radius $\gamma_s$ centered at $\bc_i$, we have $P_s = \cup_{\bc_i \in C}B_d(\bc_i, \gamma_s) \supseteq B_d(\b0,1)$. In simpler terms, we can associate with the training points a collection of balls $p_i$, each with a center $c_i$ and a radius $\gamma_s$, forming the set $P_s$. We can assign each training point to the specific ball it belongs to. If a training point belongs to more than one ball, we handle this by assigning it to the ball with the smallest distance to its center or breaking the tie arbitrarily if multiple centers exist.

For any input $\bx$ that falls within a ball $p \in P_s$, we can consider any training point $\bx_p$ that is assigned to the ball $p$ using the method mentioned earlier as supporting evidence for $\bx$ in $p$. The number of evidence points within ball $p$ can be denoted as $n_p$. The set $\{\bx_{p,i}\}_{i=1}^{n_p}$ represents the evidence points. Applying equation (\ref{uniformbound}) to all evidence points, summing up the $n_p$ inequalities, and then taking the average leads us to the following inequality:
\begin{align} \label{allboundE}
\|f(\bx) - \mathcal M(\bx) \|_2  \leq (K_f + K_{\mathcal M}) \gamma_s + \bar{\varepsilon}_p,
\end{align}
where $\bar{\varepsilon}_p$ represents the mean of the prediction errors, which is calculated as follows:
\begin{align} \label{meanE}
\bar{\varepsilon}_p = \frac{1}{n_p} \sum_{i=1}^{n_p} \| f(\bx_{p,i}) - \mathcal M(\bx_{p,i})\|_2.
\end{align}

In the given context, we consider an arbitrary probability density function $\mathcal P$, defined over the unit ball $B_d(\mathbf{0},1)$. Equation (\ref{allboundE}) can be applied to every point $\mathbf{x}$ in each region $p$ in $P_s$, and then summed over all regions in $P_s$ to obtain the following inequality:
\begin{align} \label{totalerr}
\sum_{p \in P_s} \int_p  \|f(\mathbf{x}) - \mathcal M(\mathbf{x}) \|_2 \mathcal P(\mathbf{x}) \; d\mathbf{x} & \leq (K_f + K_{\mathcal M}) \gamma_s + \sum_{p \in P_s} \int_p  \bar{\varepsilon}_p \mathcal P(\mathbf{x})\; d\mathbf{x}.
\end{align}
Here, $\bar{\varepsilon}_p$ represents a constant in region $p$ in the last equation.

We relied on the assumption of densely covering to derive Eq. (\ref{totalerr}). To represent the last term in the equation, we employed the assumption of histogram approximation. A key step in this process was the transformation of the term $\int_p \mathcal P(\bx) d\bx$ into $\frac {n_p}{N}$, where $N$ is the number of training points and $n_p$ is the number of evidence points in ball $p$. Utilizing Eq. (\ref {meanE}) for $ \bar {\varepsilon}_p$, we obtained the following expression:
\begin{align} \label{histo}
\sum_{p \in P_s}\bar{\varepsilon}_p \int_p  \mathcal P(\bx)\; d\bx & = \sum_{p \in P_s}\bar{\varepsilon}_p \frac{n_p}{N} = \frac{1}{N} \sum_{p \in P_s} \sum_{i=1}^{n_p} \|f(\bx_{p,i}) - \mathcal M(\bx_{p,i}) \|_2 \nonumber \\& = \frac{1}{N} \sum_{i=1}^N \|f(\bx_i) - \mathcal M(\bx_i) \|_2.
\end{align}
By substituting this expression into Eq. (\ref{totalerr}), we can limit the generalization error for any distribution of input data in $B_d(\b0, 1)$ using the following inequality:
\begin{align}  \label{gen2-norm}
\sum_{p \in P_s} \int_p  \|f(\bx) - \mathcal M(\bx) \|_2 \mathcal P(\bx) \; d\bx & \leq  \frac{1}{N} \sum_{i=1}^N \|f(\bx_i) - \mathcal M(\bx_i) \|_2 + (K_f + K_{\mathcal M}) \gamma_s. 
\end{align}

The first term on the right-hand side represents the empirical 2-norm loss. 
 By considering the Lipschitz constants, it is demonstrated that the difference between generalization and empirical errors can be bounded with a training data set that satisfies the mentioned assumptions. It's worth noting that there exists a bound on the minimal number of smaller balls required to form a covering of a ball of larger radius in any dimension \cite{verger2005covering}. Additionally, the smaller balls' enclosing radius $\gamma_s$ cannot be infinitely small. As emphasized in \cite{glazyrin2019covering}, it is essential for $\gamma_s$ to satisfy the inequality 
\begin{align} \label{coverN0}
N \gamma_s \geq d
\end{align}
to cover the unit $d$-ball completely. In the analysis of generalization error for Lipschitz functions with regression networks, we can establish the following bound:
\begin{citedlem}  \label{generalizationbound}
Suppose the target function $f: B_d(\b0, 1) \rightarrow \R^l$ is a Lipschitz function with Lipschitz constant $K_f$. The function $f$ is learned by regression network $\mathcal M$ using $N$ training points, $\{(\bx_i,  f(\bx_i))\}_i$.  It is assumed that the training points fulfill the condition of being $\gamma_s$-densely covering the unit $d$-ball and the assumption of the histogram approximation. For an arbitrary density function $\mathcal P$, we can bound the generalization error of learning $f$ using $\mathcal M$. The network $\mathcal M$ computes a Lipschitz function of Lipschitz constant $K_{\mathcal M}$. The bound is given by:
\begin{align} \label{rmse}
\int_{B_d(\b0,1)}  \|f(\bx) - \mathcal M(\bx) \|_2 \mathcal P(\bx) \; d\bx \leq   \frac{1}{N} \sum_{i=1}^N \|f(\bx_i) - \mathcal M(\bx_i) \|_2 +(K_f + K_{\mathcal M}) \gamma_s.
\end{align}
\qed
\end{citedlem}

\subsubsection{One-hot classification networks}

We consider the generalization bounds of classifiers that utilize one-hot coding systems widely used in network classifiers. These classifiers partition the input domain $B_d(\b0, 1)$ into a finite number of regions, where each region is assigned to a class. In contrast to the assumption of Lipschitz continuity for regression, these classifiers are piecewise constant functions, assigning inputs to a single class $i$ within the range of $\{1, \cdots, l\}$. The standard coordinate basis $\be_i$ represents class $i$ in a one-hot coding system. To compute the 0/1 error function, we can use the formula below: for any $f(\by), f(\bz) \in \{\be_1, \cdots, \be_l\}$,
\begin{align} \label{lossc}
g(f(\by), f(\bz)) = \frac{1}{\sqrt 2} \| f(\by) - f(\bz)\|_2 = 
\begin{cases}
1 \text{ if  $f(\by) \neq f(\bz)$ } \\
0 \text{ otherwise.}
\end{cases}
\end{align}
For a given classifier $f$, a closed ball exists around any point $\bx$ in the input domain, where all the points inside this ball have the same class as that of $\bx$. If $\bx$ lies on the classification boundary of $f$, the radius of the ball is zero. We use the notation $\eta_f(\bx) \geq 0$ to denote the maximum radius of the ball centered at $\bx$ such that $f(\by) = f(\bx)$ for all $\by$ belonging to $B_d(\bx, \eta_f(\bx))$. Therefore, the class of $f$ at any point $\by$ is the same as that of $\bx$ when the distance between $\bx$ and $\by$ is less than or equal to $\eta_f(\bx)$. 
Hence, we have 
\begin{align} \label{hfunction}
h_f(\bx; \| \bx - \by\|_2) = g(f(\bx), f(\by)) = 0 \text{ if $\|\bx - \by \|_2\leq \eta_f(\bx)$}.
\end{align}
The parameter $\eta_f(\bx)$ is the affirmative parameter of the function $f$ at point $\bx$. This parameter indicates the maximum radius of a ball around $\bx$ in the input domain with the same class as $\bx$. If the distance between $\bx$ and another point $\by$, denoted by $\|\bx - \by \|_2$, is greater than $\eta_f(\bx)$, the class of $\by$ cannot be determined from $h_f(\bx; \|\bx - \by\|_2)$. However, we can bound function $h_f$ with the following radius function:
\begin{align} \label{outerball}
\hat h_f(\bx; \|\bx - \by\|_2) = \begin{cases}
1 \text{ if $\| \bx - \by \|_2 > \eta_f(\bx)$} \\
0 \text{ otherwise.}
\end{cases}
\geq h_f(\bx; \| \bx - \by \|_2).
\end{align}
$\hat h_f(\bx;  \|\bx - \by\|_2)$ is a non-decreasing function of $\| \bx - \by\|_2$ for any function $f$. 
In the case where $\bx$ is on the classification boundary, $\hat h_f(\bx; \| \bx-\by\|_2)$ will always be equal to 1 for any value of $\by$.

The classification network $\mathcal M$ is responsible for learning the classifier function $f$ by utilizing $N$ training data pairs $(\bx_i, f(\bx_i))$. This is achieved through minimizing the empirical loss function $\frac{1}{N} \sum_i g(f(\bx_i), \mathcal M(\bx_i))$. The function of $\mathcal M$ is piece-wise constant, dividing the input domain into non-overlapping regions and assigning each region a specific class. The collection of all these regions is denoted as $P_s$. Each region $p$ is associated with enclosing balls, and can be linked with the smallest enclosing radius parameter $\gamma_p$. The overall smallest enclosing radius among all regions is represented as $\gamma_s$, which is equal to the maximum value of all the enclosing radius parameters $\{\gamma_p\}$ for all the regions in $P_s$.

The training points fulfill the assumption of $\gamma_s$-densely covering means that every region of $P_s$ must have at least one training point. 
For any given input point $\bx \in p \in P_s$, let $\bx_p$ be a training point in the same region of $\mathcal M$ as $\bx$. The distance between $\bx$ and $\bx_p$ will always be less than or equal to $\gamma_s$, and $\bx_p$ serves as the evidence of $\bx$ in region $p$. To calculate the error between $f(\bx)$ and $\mathcal M(\bx)$ with respect to the evidence point $\bx_p$, we can use the following equation:
\begin{align} \label{maingenc}
g(f(\bx), \mathcal M(\bx))& = \frac{1}{\sqrt 2} ( \|f(\bx) - \mathcal M(\bx) + f(\bx_p) - f(\bx_p) + \mathcal M(\bx_p) - \mathcal M(\bx_p)\|_2) \nonumber \\
& \leq \frac{1}{\sqrt 2}(\| f(\bx) - f(\bx_p)\|_2 + \| f(\bx_p) - \mathcal M(\bx_p) \|_2+\|  \mathcal M(\bx_p) - \mathcal M(\bx)\|_2)  \nonumber \\
&  \leq g(f(\bx), f(\bx_p)) + \varepsilon_p.
\end{align}
Here, we define $\frac{1}{\sqrt{2}} \| f(\bx_p) - \mathcal M(\bx_p) \|_2$ as $g(f(\bx_p), \mathcal M(\bx_p)) = \varepsilon_p$. 
As points in the same region of $P_s$ of $\mathcal M$ have the same class, the term $ \|\mathcal M(\bx_p) - \mathcal M(\bx)\|_2$ is zero. 

Denote $\{\bx_{p,i}\}_{i=1}^{n_p}$ as the set of $n_p$ evidence training points in region $p$. By applying Eq. (\ref{maingenc}) to each point of evidence in region $p$ and then computing the average, we can obtain the following result:
\begin{align} \label{maingenc1}
g(f(\bx), \mathcal M(\bx))  & \leq \frac{1}{n_p} \sum_{i=1}^{n_p} g(f(\bx), f(\bx_{p,i})) + \bar{\varepsilon}_p \nonumber\\
& \leq \max_{i} g(f(\bx), f(\bx_{p,i})) + \bar{\varepsilon}_p.
\end{align}
To assess the expected error for a given probability distribution $\mathcal P$, we use Eq. (\ref{maingenc1}) for each point $\bx$ in a region $p$, and then sum over all regions of $P_{s}$ to obtain: 
\begin{align} \label{totalerrc}
\sum_{p \in P_{s}} \int_p  g(f(\bx), \mathcal M(\bx)) \mathcal P(\bx) \; d\bx & \leq  \sum_{p \in P_{s}}\int_p  \max_{i} g(f(\bx), f(\bx_{p,i})) \mathcal P(\bx) d\bx  + \sum_{p \in P_{s}} \int_p \bar{ \varepsilon}_p \mathcal P(\bx)\; d\bx.
\end{align}
In the equation, the left side represents the generalization error, while the right side consists of two integrals. The first integral is the maximum value of the function g calculated for all evidence points $\bx_{p,i}$ and all points $\bx$ in region $p$. The second term in Eq. (\ref{totalerrc}) can be further derived using the assumption of histogram approximation. By using a similar approach as in Eq. (\ref{histo}), we can relate this term to the average error $\bar{\varepsilon}_p$ for all training points in the region p. Thus, Eq. (\ref{totalerrc}) can be expressed as:
\begin{align} \label{totalerrc1}
\sum_{p \in P_{s}} \int_p  g(f(\bx), \mathcal M(\bx)) \mathcal P(\bx) \; d\bx & \leq  
\sum_{p \in P_{s}}\int_p  \max_{i} g(f(\bx), f(\bx_{p,i})) \mathcal P(\bx) d\bx  + \frac{1}{N \sqrt{2}} \sum_i \| f(\bx_i) - \mathcal M(\bx_i) \|_2 \nonumber \\
& =  \sum_{p \in P_{s}}\int_p  \max_{i} g(f(\bx), f(\bx_{p,i})) \mathcal P(\bx) d\bx  + \frac{1}{N} \sum_i g(f(\bx_i), \mathcal M(\bx_i)).
\end{align}
We can utilize two facts to calculate an upper bound for the first term in Eq. (\ref{totalerrc}). The first is that $\| \bx - \bx_{p, i}\| \leq  \gamma_s$, and the second is that $\hat h_f$ is a non-decreasing function, as per Eqs. (\ref{outerball}) and (\ref{hfunction}). By using these facts, we obtain the following inequality:
\begin{align} \label{firstbound}
\sum_{p \in P_{s}}\int_p \max_i g(f(\bx), f(\bx_{p,i})) \mathcal P(\bx) d\bx &\leq  \sum_{p \in P_{s}}\int_p \max_i \hat h_f(\bx; \|\bx - \bx_{p,i}\|_2) \mathcal P(\bx) d\bx  \nonumber \\
&\leq \int \hat h_f(\bx; \gamma_s) \mathcal P(\bx) d\bx \nonumber \\
&\leq \int 1([\gamma_s \geq \eta_f(\bx)]) \mathcal P(\bx) d\bx.
\end{align}
Here, $1([\gamma_s \geq \eta_f(\bx)])$ is an indicator function for $\gamma_s \geq \eta_f(\bx)$, and the last inequality is a result of Eq. (\ref{outerball}). 
For a given function $f$, the value of $\eta_f(\bx)$ remains fixed. Moreover, the function $1([\gamma_s \geq \eta_f(\bx)])$ is an increasing function of $\gamma_s$. Therefore, a network $\mathcal M$ with a lower value of $\gamma_s$ can lead to a reduced generalization bound.

The probability mass that contributes to $\int 1([\gamma_s \geq \eta_f(\bx)]) \mathcal P(\bx) d\bx$ on the points in the input domain satisfying $\eta_f(\bx) \leq \gamma_s$ can be bounded. These points are within a distance no greater than $\gamma_s$ from the classification boundary of $f$, denoted by $\partial f$. By bounding the probability mass by the volume of the points over a solid $d$-tube, which is a Cartesian product of a $d-1$ ball of radius $\gamma_s$ and a curve of length $|\partial f|$, we obtain the following inequality:
\begin{align} \label{hcylinder0}
 \int 1([\gamma_s \geq \eta_f(\bx)]) \mathcal P(\bx) d\bx \leq c_{\mathcal P} \text{vol }B_{d-1}(\b0, \gamma_s) |\partial f| \leq (\gamma_s)^{d-1}\text{vol }B_{d-1}(\b0, 1) |\partial f|.
\end{align}
In the above, we utilize the fact that the volume of the $d$-tube equals $\text{vol } B_{d-1}(\b0, \gamma_s) |\partial f|$, $B_{d-1}(\b0, \gamma_s) = (\gamma_s)^{d-1} B_{d-1}(\b0, 1)$, and the constant $c_{\mathcal P}$, which depends on $\mathcal P$, is less than or equal to $1$. By substituting Eqs. (\ref{hcylinder0}) and (\ref{firstbound}) into Eq. (\ref{totalerrc1}), a generalization bound for a one-hot coding classifier can be obtained.

\begin{citedlem}  \label{generalizationboundc}
Suppose the training data meets the assumptions of $\gamma_s$-densely covering the unit $d$-ball and the histogram approximation, where $\gamma_s$ is the smallest enclosing radius of the classification regions derived by the one-hot classification network $\mathcal M: B_d(\b0,1) \rightarrow \{\be_1, \cdots, \be_l\}$. Each classification region is associated with a specific class. We can bound the generalization error of learning the classification function $f: B_d(0, 1) \rightarrow \{\be_1, \cdots,  \be_l\}$ using $\mathcal M$ with $N$ training data $\{(\bx_i,  f(\bx_i))\}_i$ as follows:
\begin{align} \label{gen2-normc}
\int_{B_d(\b0, 1)}g(f(\bx), \mathcal M(\bx))\mathcal P(\bx) \; d\bx \leq 
 \frac{1}{N} \sum_{i=1}^N g(f(\bx_i), \mathcal M(\bx_i)) +  (\gamma_s)^{d-1}\text{vol }B_{d-1}(\b0, 1) |\partial f|,
 \end{align}
where $g(f(\bx), \mathcal M(\bx))$ is the 0/1 error, and $|\partial f|$ represents the size of the classification boundary induced by $f$. The bound holds regardless of the density $\mathcal P$ over $B_d(\b0,1)$.
\qed
\end{citedlem}

\subsubsection{Network parameters}

We have discovered a link between the generalization bound and the radius of balls that encapsulate the local geometry of a function. Our next goal is to establish a connection between the radius and the number of parameters in a network. Specifically, we aim to determine the number of parameters required for a network to cover the $d$-dimensional unit balls with a desired radius of $\gamma_s$. We derive this relationship by assessing how many parameters are needed for a network to reduce a covering radius to half its size.

For a network $\mathcal N_l$ with covering balls of radius $\gamma_l$, and with $\#W_l$ parameters, we have an oracle to generate a network $\mathcal N_{l+1}$ from $\mathcal N_l$. The resulting network reduces the radius of the covering balls to $\gamma_{l+1} \leq \gamma_l/2$ and the number of parameters to $\#W_{l+1} \leq  (1 + \alpha) \#W_l$, for all $l > 0$, with $\#W_0 = 1$. The value of $\alpha \geq 1$ depends on the hypothesis class and the oracle method to reduce the enclosing radius in a network.

We can apply the oracle process recursively to decrease the enclosing radius to a desired value, $\gamma_s$. To reduce the radius from $1$ to $\gamma_s$, we can apply the procedure $L$ times with $ L = -\log_2 \gamma_s$. Let the resulting network be $\mathcal N_L$. Using  $\#W_{l+1} \leq  (1 + \alpha) \#W_l$, we can deduce the number of parameters in network $N_L$ with $\#W_L \leq (1 +\alpha)^L$.  Hence, 
$\log_2 \#W_L\leq - \log_2 (1+  \alpha) \log_2 \gamma_s $. This gives an inverse relation between $\gamma_s$ and $\#W_L$ to the power $\frac{1}{\log_2 (1 + \alpha)}$, 
with 
\begin{align}\label{gammaW}
\gamma_s \leq  \frac{1}{\#W_{L}^{1/\log_2 (1 + \alpha)}}.
\end{align}

We substitute Eq. (\ref{gammaW}) into Lemmas \ref{generalizationbound} and \ref{generalizationboundc} to obtain the generalization bounds inversely proportional to the power of the number of parameters of a network.

\begin{citedprop}   \label{prop1}
Suppose the target function $f: B_d(\mathbf{0}, 1) \rightarrow \mathbb{R}^l$ is a Lipschitz function with Lipschitz constant $K_f$. Assume the unknown function $f$ is learned by the regression network $\mathcal{M}$ using $N$ training points, $\{(\mathbf{x}_i,  f(\mathbf{x}_i))\}_i$. The training points satisfy the assumptions of $\gamma_s$-densely covering the unit $d$-ball and the histogram approximation. The network $\mathcal{M}$ computes a Lipschitz function of Lipschitz constant $K_{\mathcal{M}}$. Denote $\#W_{\mathcal{M}}$ the number of parameters of $\mathcal{M}$. For arbitrary density function $\mathcal{P}$, we can bound the generalization error with the bound given by
\begin{align} \label{rmse10}
\int_{B_d(\mathbf{0},1)}  \|f(\mathbf{x}) - \mathcal{M}(\mathbf{x}) \|_2 \mathcal{P}(\mathbf{x}) \; d\mathbf{x} \leq   \frac{1}{N} \sum_{i=1}^N \|f(\mathbf{x}_i) - \mathcal{M}(\mathbf{x}_i) \|_2 +\frac{(K_f + K_{\mathcal{M}})}{ \#W_{\mathcal{M}}^{1/\log_2 (1+\alpha)}},
\end{align}
where $\alpha \geq 1$ depends on the hypothesis class and the construction method to achieve  the desired covering of the unit $d$-ball.
\qed
\end{citedprop}

\begin{citedprop} \label{prop2}
Suppose the training points satisfy the assumptions that they $\gamma_s$-densely cover the unit $d$-ball, and the histogram approximation holds. We can bound the generalization error of learning a one-hot classification function $f$ using $\mathcal M$ with $N$ training data points $\{(\bx_i,  f(\bx_i))\}_i$ as follows:
\begin{align} \label{gen2-normc10}
\int_{B_d(\b0, 1)}g(f(\bx), \mathcal M(\bx))\mathcal P(\bx) \; d\bx \leq 
 \frac{1}{N} \sum_{i=1}^N g(f(\bx_i), \mathcal M(\bx_i)) +  \frac{\text{vol }B_{d-1}(\b0, 1) |\partial f|}{\#W_{\mathcal M}^{(d-1)/\log_2(1 + \alpha)}},
 \end{align}
Here, $g(f(\bx), \mathcal M(\bx))$ is the 0/1 error, $\#W_{\mathcal M}$ is the number of parameters of $\mathcal M$, and $|\partial f|$ represents the size of the classification boundary induced by $f$. This bound holds regardless of the density $\mathcal P$ over $B_d(\b0,1)$. The parameter $\alpha \geq 1$ depends on the hypothesis class and how to construct a network to achieve the desired covering of the unit $d$-ball.
\qed
\end{citedprop}

When comparing two methods of constructing networks to adapt to the local geometry of a function with different values of $\alpha$, our analysis indicates, for both regression and classification purposes, the method with a smaller $\alpha$ yields a lower generalization error assuming that both methods exhibit the same empirical loss. In Section \ref{oracle}, we demonstrate the oracle's implementation by creating a neural network consisting of a series of affine linear mappings and ReLU activation functions. We provide evidence that for deep neural networks as described in Eq. (\ref{series}), the validity of Propositions \ref{prop1} and \ref{prop2} holds when $\alpha \geq 1$.

\subsection{Concentration bound and sample complexity with fixed radius random balls}

Our earlier analysis reveals the local geometry parameter $\gamma_s$ and the assumptions of training data (refer to Lemmas \ref{generalizationbound} and \ref{generalizationboundc}) primarily impacts the generalization error for learning both regression and classification functions. We must carefully consider these assumptions to examine the concentration results of the generalization error bounds. Specifically, we need to address the concentration bound for the assumption that the unit $d$-ball is densely covered using random balls of a fixed radius and derive the concentration bound for the assumption of histogram approximation. Our goal is to find the concentration parameters $(\epsilon, \delta)$ based on the radius $\gamma_s$ of random balls and determine the sample complexity $m_0$, which is the minimum number of training points required to achieve the concentration result represented by $(\epsilon, \delta)$. It is essential that the concentration parameter $(\epsilon, \delta)$ and sample complexity $m_0$ remain independent of samples and use the least information from the density function $\mathcal P$.

The derivations presented below are based on the assumption that we can ensure a minimum probability mass for any ball of radius $\gamma_s$, with its center inside the unit $d$-ball. We refer to this  assumption as the "uniform probability lower bound," represented as:
\begin{align} \label{lowerkappa}
P(\bx; B_d(\bx, \gamma_s)) = \int_{\bz \in B_d(\bx, \gamma_s)} \mathcal P(\bz) d\bz \geq c\gamma_s^d  > 0.
\end{align}
We can interpret $c \gamma_s^d$ as a statement similar to the mean value theorem of $P(\bx; B_d(\bx, \gamma_s))$, with the distinction that the parameter $c$ is a constant independent of $\gamma_s$ and $\mathcal P$.

\subsubsection{Covering the unit $d$-ball randomly for the first time}

We use a process denoted as $X$ to generate centers for small balls of radius $\gamma_s$ one at a time. These centers are drawn independently based on the density function $\mathcal P$ over the unit $d$-ball. The goal is to estimate the expected minimum number of random points (denoted by $\mathcal E(X_{min})$) of $X$ needed to cover the unit $d$-ball. In simpler terms, we are trying to estimate the average number of samples $m$ drawn independently according to $\mathcal P$ needed to cover the unit $d$-ball for the first time.

We define a parameter $\delta_1$ (where $0 < \delta_1 < 1$) to denote the probability that the initial $m$ samples generated by $X$ cannot cover the unit $d$-ball. Leveraging the Markov inequality, we can derive an upper bound for this probability as follows:
\begin{align} \label{coveringP}
\delta_1(m) =  P(X_{min} > m) \leq \frac{\mathcal E(X_{min})}{m}.
\end{align}
As illustrated in Lemma \ref{firstime}, it is feasible to express the value of $\mathcal E(X_{min})$ using the uniform probability lower bound assumption with the following equation:
\begin{align} \label{kappavalue}
\mathcal E(X_{min}) \leq \frac{1}{c \gamma_s^d}.
\end{align}
By examining Eqs. (\ref{coveringP}) and (\ref{kappavalue}), we can conclude that the parameter $\delta_1$ and the sample size $m$ must adhere to the following relationship:
\begin{align} \label{coveringdelta}
\delta_1(m) \leq \frac{1}{c m \gamma_s^d}.
\end{align}
The probability mass $\delta_1(m)$ signifies the chance that the initial $m$ samples of the random process $X$ do not cover $B_d(\b0, 1)$. The complement $1-\delta_1(m)$ denotes the probability of the initial $m$ samples of $X$ forming a covering. Subsequent sections will delve into distinct developments for regression and classification functions.

\subsubsection{Regression functions}

When we focus solely on the assumption of densely covering, we can establish the following limitation for Eq. (\ref{totalerr}):
\begin{align} \label{rmseRB-1}
P[\int_{B_d(\b0,1)} \|f(\bx) - \mathcal M(\bx) \|_2 \mathcal P(\bx) \; d\bx - \sum_{p \in P_s}  \bar\varepsilon_p \int_p \mathcal P(\bx) d\bx    \leq (K_f + K_{\mathcal M}) \gamma_s] > 1-\delta_1(m).
\end{align}
This limitation outlines the conditions under which applying the densely covering assumption to generalization error is valid. The likelihood $1 - \delta_1(m)$ represents the probability that the first $m$ balls, with centers sampled according to the density function $\mathcal P$, will cover the input domain.

We will use a training set $S$ containing $m$ ordered pairs $(\bx_i, f(\bx_i))$ to denote an instance of the covering process $X$. The order in which training points in $S$ were drawn by $X$ according to probability $\mathcal P$ is denoted by the index $i$. We suppose that $S$ forms a covering of the unit $d$-ball with smaller balls of radius $\gamma_s$, and denote $X_{min}(S)$ the minimum number of points in $S$ to cover the unit $d$-ball for the first time. The fact that $S$ covers the unit $d$-ball and $|S| =m$ implies that $X_{min}(S) \leq m$.

To investigate the assumption of histogram approximation, we can analyze the distribution of points in $S$ within each covering ball. Let us recall that $P_s$ represents the set of covering balls with radius $\gamma_s$, and each ball is associated with a bin in the histogram that approximates the probability function $\mathcal P$.
To calculate the mass concentration for each bin $p$ in $P_s$, we define a random variable $\tilde Z_p^m$ to represent the histogram mass over bin $p$. We also identify the center of the bin as $\bp_c$. There are $|P_s|$ bins, each with a volume of $\text{vol }B_d(\bp_c, \gamma_s) = \gamma_s^d \text{vol }B_d(\b0, 1)$.  Denote an observation sequence of $m$ balls as $X_1, \cdots, X_m$. The histogram mass over bin $p$ is the average observation number in the sequence within the bin, given by
\begin{align} \label{randomZ}
\tilde Z_p^m =\frac{1}{m} \sum_{i=1}^m I(X_i \in p) = \frac{n_p}{m}.
\end{align}
$I$ is the indicator function that takes the value $1$ or $0$, and $n_p$ is the number of observations in bin $p$. It is important to note that
\begin{align}
\mathcal E\{\tilde Z_p^m\} = \int_{B_d(\bp_c, \gamma_s)} \mathcal P(\bx) d\bx.
\end{align}

When applying the Hoeffding's inequality to the sum of $m$ independent random variables, each taking value in the range $[0, \frac{1}{m}]$ and represented as 
\begin{align}
\frac{1}{m} I(X_1 \in p), \cdots, \frac{1}{m} I(X_m \in p),
\end{align}
 for all $t_p > 0$, we obtain the inequality:
\begin{align} \label{Hoeffp}
P( \int_p \mathcal P(\bx) d\bx-  \frac{1}{m} \sum_{i=1}^m I(X_i \in p) \geq   t_p)  \leq  \exp(-2mt^2).
\end{align}

In the following, we utilize the information from Eq. (\ref{meanE}) and (\ref{randomZ}) to derive the expression:
\begin{align} \label{secondterm}
\frac{\bar{\epsilon}_p  \sum_{i=1}^m I(X_i \in p)}{m} = \frac{1}{m}\sum_{i=1}^{n_p} \| f(\bx_{p,i} - \mathcal M(\bx_{p, i}) \|_2.
\end{align}

In order to utilize Hoeffding's inequality for region $p$, we can set the precision parameter $t_p$ as $\frac{k}{|P_s| \bar \epsilon_p}$. By substituting Eq. (\ref{secondterm}) into Eq. (\ref{Hoeffp}), and using the assumption on $S$ with $X_{min}(S) = |P_s| \leq m$, we can represent the inequality as:
\begin{align} \label{Hoeffp1}
P( \bar \epsilon_p \int_p \mathcal P(\bx) d\bx- \frac{1}{m} \sum_{i=1}^{n_p} \| f(\bx_{p,i}) - \mathcal M(\bx_{p,i})\|_2 \geq  k/|P_s|)  \leq  \exp(- \frac{2mk^2}{ |P_s|^2 \bar\epsilon_p^2}) \leq \exp(- \frac{2k^2}{ |P_s| \bar\epsilon_p^2}).
\end{align}
To simplify, we can assume that the range of the regression loss $\|f(\bx_i) - \mathcal M(\bx_i)\|_2$ is within the interval [0, 1]. Consequently, $\bar \epsilon_p$ is also within the range [0, 1], and $\exp(- \frac{2k^2}{ |P_s| \bar\epsilon_p^2}) \leq  \exp(- \frac{2k^2}{ |P_s|})$. Thus, the upper bound is independent of $p$ and Eq. (\ref{Hoeffp1}) is expressed as 
\begin{align} \label{Hoeffp15}
P( \bar \epsilon_p \int_p \mathcal P(\bx) d\bx- \frac{1}{m} \sum_{i=1}^{n_p} \| f(\bx_{p,i}) - \mathcal M(\bx_{p,i})\|_2 \geq  k/|P_s|)  \leq  \exp(- \frac{2k^2}{ |P_s|}) 
\end{align}

Each training point is used only once in our setting, and the concentration bound for $p$, as represented in Eq. (\ref{Hoeffp15}), is independent of the other regions in $P_s$. By considering the conjunction of the concentration bound for all $p \in P_s$, we can derive the following concentration inequality for $P_s$:
\begin{align} \label{Hoeffp2}
P(\sum_{p \in P_s} \bar \epsilon_p\int_p \mathcal P(\bx) d\bx- \frac{1}{m}\sum_{i=1}^m \| f(\bx_i) - \mathcal M(\bx_i)\|_2 \geq  k)  \leq  \prod_{p \in P_s} \exp(- \frac{2k^2}{|P_s|}) = \exp(- 2k^2).
\end{align}
Let us consider the inequality where $k = (K_f + K_{\mathcal M}) \gamma_s$, which is equivalent to:
\begin{align} \label{meanreg1r}
P(\sum_{p \in P_s} \bar \epsilon_p\int_p \mathcal P(\bx) d\bx- \frac{1}{m}\sum_{i=1}^m \| f(\bx_i) - \mathcal M(\bx_i)\|_2 <  (K_f + K_{\mathcal M}) \gamma_s)  > 1 - \exp(- 2(K_f + K_{\mathcal M})^2 \gamma_s^2)
\end{align}

The probability of obtaining the concentration inequality for the assumptions of densely covering and histogram approximation is the combined likelihood of two conditions. Firstly, it is the likelihood that the set $S$ of $m$ random balls forms a covering of the unit $d$-ball, expressed as $1-\delta_1(m)$ in Eq. (\ref{rmseRB-1}). Secondly, it is the likelihood that the balls in $S$ satisfy the Hoeffding's inequalities as expressed in inequality (\ref{meanreg1r}). To combine the two likelihoods, we introduce a precise parameter $\epsilon$, which can take values within the range of $(0, 1)$, and utilize the equation shown below to determine the value of $\epsilon$:
\begin{align} \label{coveringepsilonreg}
\epsilon = 2 (K_f + K_{\mathcal M}) \gamma_s.
\end{align}
By ensuring that $\epsilon$ is less than 1, we can obtain a bound on enclosing radius with 
\begin{align} \label{gammareglow}
\gamma_s < \frac{1}{2(K_f + K_{\mathcal M})}.
\end{align}

The probability of confidence for $S$ to satisfy statements of Eqs.  (\ref{rmseRB-1}) and (\ref{meanreg1r}) using $\epsilon$ is $(1- \exp(- \frac{\epsilon^2}{2})) (1- \delta_1(m)) =  1 - \delta(S)$, where $\delta(S)$ is a sample-dependent confidence parameter, with 
\begin{align} \label{coveringdeltareg}
\delta(S) =  \delta_1(m) + (1 - \delta_1(m))  \exp(- \frac{\epsilon^2}{2}).
\end{align}
We can establish a uniform bound for the confidence parameter, denoted as $\delta$, independent of $S$ to achieve the concentration inequality for the assumptions of densely covering and histogram approximation. 
Using the bound for $\delta_1(m)$ in Eq. (\ref{coveringdelta}), we can obtain  
\begin{align} \label{coveringdeltareg1}
\delta(S)  < \frac{1}{c m \gamma_s^d} +  \exp(- \frac{\epsilon^2}{2}) \leq \delta < 1. 
\end{align}
This definition of $\delta$ gives a constraint on $\delta$ and $\epsilon$ with 
\begin{align} \label{deltaepsilon}
\delta > \exp(- \frac{\epsilon^2}{2}).
\end{align}
The requirement of $\delta < 1$ provides a constraint on $m$ with
\begin{align} \label{coveringm0reg}
m > m_0= \frac{1}{c \gamma_s^d (\delta -\exp(-\frac{\epsilon^2}{2}))} >\frac{1}{c \gamma_s^d (1 -\exp(-\frac{\epsilon^2}{2}))}.
\end{align}

The values $\epsilon$, $\delta$, and $m_0$ obtained from $m$, $\gamma_s$, and their constraints allow us to establish the concentration bound for regression functions.

\begin{citedthm}  \label{generalizationboundRB}
Let $f: B_d(\mathbf{0}, 1) \rightarrow \mathbb{R}^l$ be a Lipschitz function with Lipschitz constant $K_f$, and $\mathcal{M}$ be a regression network that learns $f$. The Lipschitz constant of $\mathcal{M}$ is $K_{\mathcal{M}}$. We set $\epsilon = 2 (K_f + K_{\mathcal{M}}) \gamma_s < 1$, and assume a lower bound on probability mass according to Eq. (\ref{lowerkappa}). With $m$ samples, $\{(\mathbf{x}_i, f(\mathbf{x}_i))\}_i$, we train $\mathcal{M}$ to achieve $\|f(\mathbf{x}_i) - \mathcal{M}(\mathbf{x}_i) \|_2 \in [0, 1]$ for all $i$. \\
(i)  By choosing appropriate values for $\delta$ and $m_0$ using Eqs. (\ref{coveringdeltareg1}), (\ref{deltaepsilon}), and (\ref{coveringm0reg}), we can derive the generalization bound of $\mathcal M$ for all $m > m_0$ as follows:
\begin{align} \label{rmseRB}
P\left[\int_{B_d(\mathbf{0},1)}  \left\|f(\mathbf{x}) - \mathcal M(\mathbf{x}) \right\|_2 \mathcal P(\mathbf{x}) \; d\mathbf{x} -  \frac{1}{m} \sum_{i=1}^m \left\|f(\mathbf{x}_i) - \mathcal M(\mathbf{x}_i) \right\|_2  < \epsilon\right] > 1-\delta,
\end{align}
where $\{\mathbf{x}_i\}$ are independent samples drawn from the density function $\mathcal P$. \\
(ii) If we replace $\gamma_s$ with the number of parameters of $\mathcal M$ using Eq. (\ref{gammaW}), we can express the above generalization bound as:
\begin{align}
\int_{B_d(\b0,1)}  \|f(\bx) - \mathcal M(\bx) \|_2 \mathcal P(\bx) \; d\bx \leq \frac{1}{m} \sum_{i=1}^m \|f(\bx_i) - \mathcal M(\bx_i) \|_2 +   \frac{2(K_f + K_{\mathcal M})}{ \#W_{\mathcal M}^{1/\log_2 (1+\alpha)}}.
\end{align}
The hypothesis class and learning algorithms determine the value of $\alpha$. For deep neural networks described in Eq. (\ref{series}), we obtain $\alpha \geq 1$.
\qed
\end{citedthm}

The concentration analysis examines the connections between the parameters $\epsilon$, $\delta$, $m_0$, $m$, and any prior knowledge. Sample complexity determines the amount of data samples needed to achieve a specific level of learning performance, as defined by $(\epsilon, \delta)$. It evaluates the effectiveness of a learning algorithm. A more effective algorithm requires fewer samples to achieve the desired performance level, reducing the data collection and storage resources. Our analysis shows that by utilizing $K_f$ and setting $K_{\mathcal M} = K_f$, we can express the sample complexity as ${\cal O}(\frac{1}{\epsilon^d \delta})$, where $d$ represents the dimension of the input domain.

\begin{citedcor} \label{samplecomplexityR}
If Theorem \ref{generalizationboundRB} holds and the Lipschitz constant $K_f$ of the target regression function $f$ is known, and if the Lipschtiz constant $K_{\mathcal M}$ of regression network $\mathcal M$ is also $K_f$, then the sample complexity for $\epsilon$ and $\delta$ is given by:
\begin{align} \label{regressionm0}
m_0 = \frac{4^d (K_f)^d}{c \epsilon^d (\delta - \exp(- \frac{\epsilon^2}{2}))}. 
\end{align}
The constant $c$ is defined in Eq. (\ref{lowerkappa}). In terms of order, the sample complexity is ${\cal O}(\frac{1}{\epsilon^d \delta})$.
\qed
\end{citedcor}

\subsubsection{Classification functions}

In addition to the regression function, we can combine the concentration bounds of the densely covering assumptions and the histogram approximation by deriving separate concentration bounds for classification functions. 

Precisely, we can follow the approach of Eq. (\ref{rmseRB-1}) to use $\sum_{p \in P_{s}} \int_p  g(f(\bx), \mathcal M(\bx)) \mathcal P(\bx) \; d\bx$ instead of $\int_{B_d(\b0,1)}  \|f(\bx) - \mathcal M(\bx) \|_2 \mathcal P(\bx) \; d\bx$ to obtain the concentration bound of the $\gamma_s$-densely covering. Moreover, we can follow the approach detailed to obtain Eq. (\ref{Hoeffp2}) and utilize $\frac{1}{m}\sum_{i=1}^m g(f(\bx_i), \mathcal M(\bx_i))$ instead of $\frac{1}{m}\sum_i \| f(\bx_i) - \mathcal M(\bx_i)\|_2$ and set $k = (\gamma_s)^{d-1}\text{vol }B_{d-1}(\b0, 1) |\partial f| $ to obtain the concentration bound of the histogram approximation.

We are interested in the probability of achieving a precision bound of $\epsilon$, which depends on two main factors. Firstly, it relies on the likelihood that $S$ forms a covering of the unit $d$-ball, denoted as $1-\delta_1(m)$. Secondly, it depends on the probability that $S$ satisfies Hoeffding's inequalities with a specific $k$ setting, with a bound given by $1-\exp(-2k^2)$ according to Eq. (\ref{Hoeffp2}). 

We can calculate the concentration results by determining the precision parameter $\epsilon$ using the equation:
\begin{align}\label{coveringepsilonregc} \epsilon = 2(\gamma_s)^{d-1}\text{vol }B_{d-1}(\b0, 1) |\partial f| \end{align}
To ensure that $\epsilon$ is less than $1$, we can establish a bound on the enclosing radius using the inequality:
\begin{align} \label{gammaclalow} \gamma_s < \left(\frac{1}{2\text{vol }B_{d-1}(\b0, 1) |\partial f|}\right)^{\frac{1}{d-1}}. \end{align}
We can then derive a uniform confidence parameter $\delta$ and its constraint, shown respectively in Eqs.  (\ref{coveringdeltareg1}) and (\ref{deltaepsilon}). Additionally, we can determine the sample complexity, denoted as $m_0$, as shown in Eq. (\ref{coveringm0reg}). In the equations, the value of $\epsilon$ is given by Eq. (\ref{coveringepsilonregc}).
The values $\epsilon$, $\delta$, and $m_0$ obtained from $m$, $\gamma_s$, and their constraints allow us to establish the concentration bound for classification functions.

\begin{citedthm}  \label{generalizationboundCB}
Suppose $f: B_d(\b0, 1) \rightarrow \{\be_1, \cdots, \be_l\}$ is a target classification function and $\mathcal M$ is a one-hot classification network that learns $f$ using $m$ training data $\{(\bx_i, f(\bx_i))\}_i$, where $\bx_i$ is the center of a ball of radius $\gamma_s$ and i.i.d. samples from the density function $\mathcal P$.
We set $\epsilon = 2(\gamma_s)^{d-1}\text{vol }B_{d-1}(\b0, 1) |\partial f| < 1$, and assume a lower bound on probability mass according to Eq. (\ref{lowerkappa}).   \\
(i)  By choosing appropriate values for $\delta$ and $m_0$ using Eqs. (\ref{coveringdeltareg1}), (\ref{deltaepsilon}), and (\ref{coveringm0reg}), we can derive the generalization bound of $\mathcal M$ for all $m > m_0$ as follows:
\begin{align} \label{gen2-normCB}
P[\int_{B_d(\b0, 1)}g(f(\bx), \mathcal M(\bx))\mathcal P(\bx) \; d\bx - 
 \frac{1}{m} \sum_{i=1}^m g(f(\bx_i), \mathcal M(\bx_i)) < \epsilon] > 1-\delta.
\end{align}
(ii) If we replace $\gamma_s$ with the number of parameters of $\mathcal M$ using Eq. (\ref{gammaW}), we can express the above generalization bound as:
\begin{align}
\int_{B_d(\b0,1)}  g(f(\bx), \mathcal M(\bx)) \mathcal P(\bx) \; d\bx \leq \frac{1}{m} \sum_{i=1}^m g(f(\bx_i), \mathcal M(\bx_i))+  \frac{\text{vol }B_{d-1}(\b0, 1) |\partial f|}{\#W_{\mathcal M}^{(d-1)/\log_2(1 + \alpha)}}.
\end{align}
The hypothesis class and learning algorithms determine the value of $\alpha$. For deep neural networks described in Eq. (\ref{series}), we obtain $\alpha \geq 1$. \\
\qed
\end{citedthm}
The following demonstrates that the sample complexity of a given function $f$ is in the order of ${\cal O}\left(\frac{1}{{\epsilon}^{\frac{d}{d-1}} \delta}\right)$.
\begin{citedcor} \label{samplecomplexityC}
Suppose $f: B_d(\b0, 1) \rightarrow \{\be_1, \cdots, \be_l\}$ is a target classification function and $\mathcal M$ is a one-hot classification network that learns $f$ using $m$ training data $\{(\bx_i, f(\bx_i))\}_i$, where $\bx_i$ are i.i.d. samples from the density function $\mathcal P$.
If Theorem \ref{generalizationboundCB} holds, then the sample complexity for $\epsilon$ and $\delta$ is given by:
\begin{align} \label{classificationm0}
m_0 =\frac{ (2 \text{vol }B_{d-1}(\b0, 1) |\partial f|)^{\frac{d}{d-1}}}{c \epsilon^{\frac{d}{d-1}} (\delta - \exp(- \frac{\epsilon^2}{2}))}.
\end{align}
The constant $c$ is defined in Eq. (\ref{lowerkappa}). In terms of order, the sample complexity is ${\cal O}\left(\frac{1}{{\epsilon}^{\frac{d}{d-1}} \delta}\right)$.
\qed
\end{citedcor}

\subsubsection{The uniform probability lower bound} \label{uniformassumption}

We drew inspiration from the idea of random covering, with foundational insights available in Hall's 1988 book \cite{hall1988introduction} and references to more recent developments provided in \cite{penrose2023random, aldous2022covering}. 
Our objective was to validate the accuracy of equation (\ref{kappavalue}), which calculates the average number of random balls with a radius $\gamma_s$ needed to cover the unit $d$-ball for the first time completely. This is done using process $X$, which generates a sequence of balls by employing a density function to select each smaller ball's center independently.

\begin{citedlem} \label{firstime}
Assume the uniform probability lower bound assumption holds. 
The average number of random balls with a radius $\gamma_s$ to cover for the first time the unit $d$-ball by process $X$ is
\begin{align}
\mathcal E(X_{min}) \leq \frac{1}{c\gamma_s^d}.
\end{align}
where $c$ is the uniform probability lower bound parameter, Eq. (\ref{lowerkappa}).
\end{citedlem}
\proof

Consider a point $\bx$ in the ball $B_d(\b0,1)$. If the center of any ball with a radius $\gamma_s$ lies outside the ball $B_d(\bx, \gamma_s)$, then the point $\bx$ cannot be contained in that ball. Therefore, the probability of a randomly chosen ball not containing $\bx$ is $1 - P(\bx; B_d(\bx, \gamma_s))$, where $P(\bx; B_d(\bx, \gamma_s))$ represents the probability of selecting a ball centered in $B_d(\bx, \gamma_s)$. 

To calculate $P(\bx; B_d(\bx, \gamma_s))$, we can evaluate the integral $\int_{\bz \in B_d(\bx, \gamma_s)} \mathcal P(\bz) d\bz$, where $\mathcal P$ is the probability density function. The probability of $\bx$ being covered by the $l$-th random ball but not by the first $l-1$ balls by process $X$ is given by $P(\bx; B_d(\bx, \gamma_s)) (1- P(\bx; B_d(\bx, \gamma_s) ))^{l-1}$. Hence, the expected number of random balls required to cover $\bx$ for the first time is
\begin{align} \label{pointlength}
{\bar l}(\bx) & = \sum_{l=1}^{\infty} l P(\bx; B_d(\bx, \gamma_s)) (1- P(\bx; B_d(\bx, \gamma_s)) )^{l-1} \nonumber \\
& = P(\bx; B_d(\bx, \gamma_s))  \sum_{l=1}^{\infty} l (1- P(\bx; B_d(\bx, \gamma_s)) )^{l-1} \nonumber \\
& = P(\bx; B_d(\bx, \gamma_s))  \frac{d\sum_{l=1}^{\infty} [1 - P(\bx; B_d(\bx, \gamma_s) )]^l}{d[1- P(\bx; B_d(\bx, \gamma_s) )]} \nonumber \\
& = P(\bx; B_d(\bx, \gamma_s))  \frac{d}{d[1- P(\bx; B_d(\bx, \gamma_s) )]}[\sum_{l=0}^{\infty}(1 - P(\bx; B_d(\bx, \gamma_s) ))^l - 1] \nonumber \\
& = \frac{1}{P(\bx; B_d(\bx, \gamma_s))} \leq \frac{1}{c \gamma_s^d}.
\end{align}
In the above calculation, we utilized the formula $\sum_{i=0}^{\infty} x^i = \frac{1}{1-x}$ for $x \in (0, 1)$ and used $dx$ to signify the derivative of $x$. 

The mean of the first time to cover the unit $d$-ball by random balls is the integral of the product of the coverage length function and the probability density function over the unit $d$-ball, i.e., 
\begin{align} \label{meancovering}
\mathcal E(X_{min})  = \int_{\bx \in B_d(\b0,1)} \bar l(\bx) \mathcal P(\bx) d\bx \leq \frac{1}{c \gamma_s^d}.
\end{align}

\qed

When working with the input probability $\mathcal P$, it is evident that dedicating sampling time to covering a less probable region is significant. The upper bound derived from Eq. (\ref{meancovering}) is quite conservative as it assumes a low probability mass in the vicinity of any $\bx$. This conservative bound fails to account for the uneven probability distribution across different areas. If one location has a low probability mass, other regions should have a higher probability mass. In line with the discussion above, whitening images is a pre-processing technique that can accelerate learning speed to convergence. This approach eliminates correlated components in input images and establishes an equalized distribution in all directions \cite{ioffe2015batch,maharana2022review}.
If we ignore boundary effects, a uniform input probability distribution leads to 
$\mathcal E(X_{min}) =\frac{1}{\gamma_s^d}$ with the following probability density function:
\begin{align}
P(\bx; B_d(\bx, \gamma_s)) = \frac{\text{vol }B_d(\b0, \gamma_s)}{\text{vol }B_d(\b0,1)} = \gamma_s^d.
\end{align}

The term $\gamma_s^d$ in Eq. (\ref{meancovering}) cannot be improved. Section \ref{expectednumber} reinforces the assertion that $\gamma_s^d$ is necessary by considering a continuous density function $\mathcal P$.

\section{Oracle implementation using DNNs} \label{oracle}

We are currently exploring methods to derive functions within the hypothesis space to achieve the oracle process that reduces the enclosing radius of the balls by at least half of their original sizes. Our focus is on investigating a deep neural network (DNN) with $L$ layers, which is structured as follows:

\begin{align} \label{series}
\mathcal M_L^0 =[\rho M_L] \circ [\rho M_{L-1}] \circ \cdots [\rho M_1].
\end{align}
Here, $M_l$ represents an affine linear mapping, and $\rho$ denotes the Rectified Linear Unit (ReLU) activation functions. The function represented by $\mathcal M_L^0$ is considered to be piece-wise continuous. Pascanu et al. \cite{pascanu2013number} previously discussed its properties and its capability for function approximations, particularly focusing on the number of regions generated for function approximation. In a related analysis, Hwang et al. \cite{hwang2019rectifying} further explored the work of Pascanu et al., with a specific emphasis on the stability of input perturbation through Lipschitz analysis.

We denote $\mathcal M_l^0 = [\rho M_l] \cdots [\rho M_1]$ as the first $l$ layers of $\mathcal M^0_L$. If $P_l^0$ is the finest input domain partition of $\mathcal M_l^0$, where each partitioning region of $P_l^0$ contains no sub-regions, then $|P_l^0|$ represents the size of $P_l^0$. When considering $P^0_l$, each partitioning region creates a polytope, an affine linear mapping domain. Additionally, we note that $P^0_{l+1} \subseteq P^0_l$, meaning that $P^0_{l+1}$ is a more detailed version of $P^0_l$ (this means every region in $P^0_{l+1}$ is within a region in $P^0_l)$. This refinement of partitioning regions in $P_l^0$ sets up a tree-like partitioning of $P_l^0$, where each region in $P^0_{l+1}$ has a unique parent region in $P^0_l$. Remindingly, this tree-like refinement of partitions is specific to $\mathcal M^0_L$ and is not a general property for networks \cite{hwang2023analysis}. 

To distinguish between the purposes of regression and classification for network $\mathcal M_L^0$, we need to incorporate the output layer into $\mathcal M_L^0$. In regression, the output layer is added at the end using an affine linear mapping $M_{L+1}$. For classification, $\mathcal M_L^0$ is augmented with $\sigma M_{L+1}$, where $\sigma$ denotes the softmax function. We note that the added affine linear and softmax functions do not contribute to the domain partitioning of $\mathcal M_L^0$. The step of assigning classes involving selecting one class from the output of $\sigma M_{L+1} \mathcal M_L^0$ is deliberately omitted. This step would involve further refining the partition $P_L^0$, but we will not delve into that as it does not change any conclusions in this context. Hence, when it comes to input domain partitioning, there is no necessity to distinguish between regression and classification purposes for network $\mathcal M_L^0$.

In the realm of $P_L^0$, every partitioning region forms a polytope, essentially the intersection of half-planes. The "diameter" of a region denotes the longest distance between two points within the region. The maximum diameter among all polytopes in a partition is equivalent to the diameter of the smallest enclosing ball of the partition. Calculating the diameter of a polytope can be intricate when dealing with polytopes of arbitrary dimension and facet number. Here, the facet of a $d$-polytope refers to the intersection of $d-1$ supporting hyperplanes of the polytope. According to Frieze and Teng \cite{frieze1994complexity}, this calculation falls into the category of problems that are both NP and co-NP hard. However, with a fixed dimension,  the diameter can be computed using a polynomial time algorithm \cite{barnette1974upper, frieze1994complexity}.

We are developing an algorithm to partition a polytope and reduce the diameters of the resulting sub-polytopes to a constant fraction of the original value. We plan to use hyperplanes passing through anchor points to make multiple cuts in the original polytope and create sub-polytopes. The plan's goal is to ensure that the diameter of the refined partition is at most half the diameter of the original polytope by applying the algorithm to every polytope in $P_L^0$ and creating a refined partition. The algorithm first divides the root node into two sub-polytopes. Then, the same method is applied to each sub-polytope until the diameters of all sub-polytopes meet the required size. As a result, each polytope in $P_L^0$ acts as the root of a tree-like structure, with intermediate and leaf nodes representing sub-polytopes. The sub-polytopes in the leaf nodes will have a diameter smaller than half of that at the root. Therefore, the collection of sub-polytopes at the leaf nodes of all trees will form a refined partition of $P_L^0$, and the diameter of the refined partition will be smaller than half that of $P_L^0$.

\begin{figure*}[tp]
\centering
\includegraphics[width=12cm]{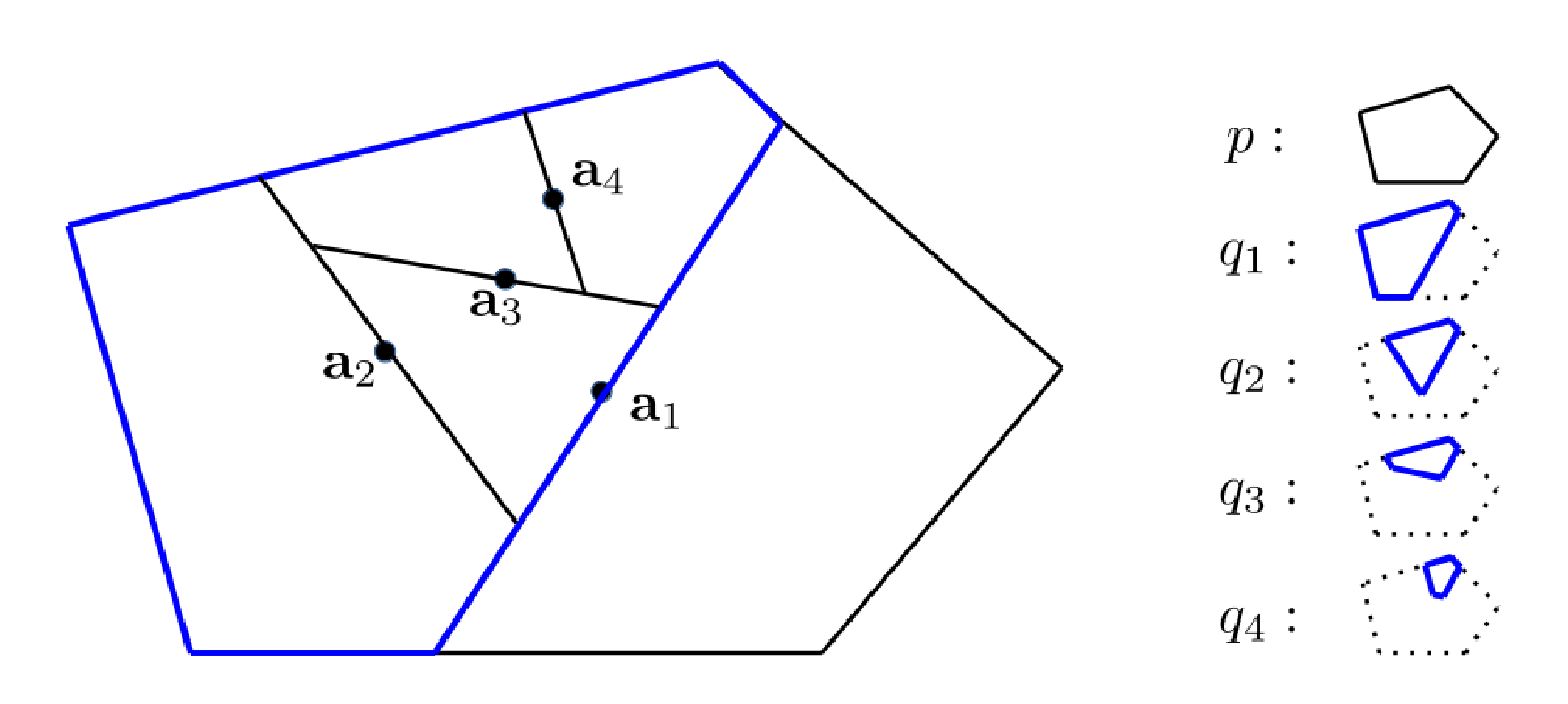}
\caption{
The convex polygon $p$ can be partitioned into sub-polytopes using sequences of hyperplane cuttings, leading to a binary tree structure with $p$ at the root. The root $p$ is divided into two sub-polygons by a hyperplane passing through the anchor point $\ba_1$, the center of the maximal inscribed ball of $p$. The sub-polygon $q_1$, colored in blue, is further divided using anchor points $\ba_2$, $\ba_3$, and $\ba_4$, corresponding to the centers of sub-polygons $q_2$, $q_3$, and $q_4$. It is worth noting that the volume of any sub-polygon retains a constant fraction of its parent polygon, as defined by Eq. (\ref{fractionvol}).}
\label{Pparfig}
\end{figure*}

The anchor point will be the center of the maximum inscribed ellipsoid in a polytope. We can calculate this point using convex optimization methods \cite{boyd2004convex}. Furthermore, we can use the property that approximates the polytope volume from inside and outside at the anchor point, as described in \cite{grotschel2012geometric, boyd2007localization}.
Let $\ba^p \in \R^d$ be the anchor point of polytope $p$. Any hyperplane passing through $\ba^p$ divides polytope $p$ into two sub-polytopes, with $q$ being one of them. The volume-reduction property shows a reduction in volume between $p$ and $q$, as demonstrated by the inequality
\begin{align}\label{fractionvol}
 \text{vol}(q) \leq \left(1-\frac{1}{d}\right) \text{vol}(p).
 \end{align}
Figure \ref{Pparfig} provides a schematic description of the tree and the cuttings. We assume that all eigenvalues of the largest inscribed ellipsoids of $p$ and subsequent sub-polytope $q$ are greater than zero, and the ratio of the smallest radius to the largest radius of a maximum inscribed ellipsoid is not less than $\zeta$. This requirement guarantees that a child sub-polytope has a non-empty interior.

\begin{figure*}[tp]
\centering
{ \includegraphics[width=5cm, height=5cm]{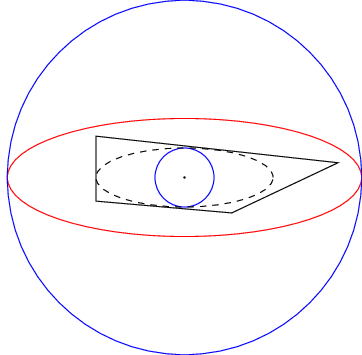}}
\caption{The dashed ellipsoid represents the maximum inscribed ellipsoid of the polygon. As per Eq. (\ref{approxeff}), the primary axis of the ellipsoid has a radius of $\eta$, while the minor axis has a radius of $\tilde \eta$, and their ratio is given by $\tilde \eta/\eta = \zeta$. The center of the ellipsoid corresponds to the highlighted anchor point of the polygon. The outer ellipsoid, shown in red and expanded by a factor of two at the boundary of the maximum inscribed ellipsoid, encloses the polygon. Enclosed within the polygon is a blue inner circle with a radius of $\tilde \eta$, and the polygon itself is enclosed by a blue outer circle with a radius of $2 \eta$. 
}
\label{fig:maximalinscribed}
\end{figure*}

Additionally, if polytope $p$ is convex (or its convex hull can be used if it's not), then the volume of polytope $p$ can be effectively approximated from the outside by enlarging a constant factor of the inner maximum inscribed ball; specifically, the outer ellipsoid of $p$ represents the boundary of the inner ellipsoid, expanded by a factor $d$ around the anchor point. 
Thus, if $\eta$ represents the radius of the primary principle axis of the inner maximum volume inscribed ellipsoid of $p$ and $\tilde \eta$ represents the smallest radius of the ellipsoid, then $d\eta$ corresponds to the radius of an enclosing ellipsoid of $p$, as expressed by the inequality:
\begin{align}  \label{approxeff}
\zeta^d \eta^d \text{vol }B_d(\b0,1) \leq \tilde \eta^d \text{vol }B_d(\b0,1)  \leq \text{vol}(p) \leq (d\eta)^d \text{vol }B_d(\b0,1),
\end{align}
where the first term uses the assumption $\tilde \eta/\eta \geq\zeta$. Figure \ref{fig:maximalinscribed} provides a schematic explanation of Eq. (\ref{approxeff}).

Now, let's consider any polytopes formed by making cuts in $p$ with a hyperplane. These new polytopes can be seen as "children" of the original polytope $p$ in a binary tree. This process can be repeated for each sub-polytope, creating a binary tree of depth $k$, with $2^k$ leaves, each leaf sub-polytope connected to a path rooted at $p$. We can label the polytope in a path at depth one as $q_1$, the polytope in the path at depth two as $q_2$, and so on. As we increase the depth $k$, the volume of $q_k$ decreases to zero, as shown in Eq. (\ref{fractionvol}). Let's denote $\eta_k$ and $\tilde \eta_k$ as the largest and smallest radii of the principal axes, respectively, of the maximum volume inscribed ellipsoid of $q_k$. Using Eqs. (\ref{fractionvol}) and (\ref{approxeff}), we obtain the inequality 
\begin{align} \label{smallrad}
\zeta^d \eta_k^d\text{vol }B_d(\b0,1) \leq \text{vol}(q_k) \leq (1-\frac{1}{d})^k \text{vol }(p).
\end{align}
This implies that the sequence $\{\eta_k\}$ decreases to zero as $k$ increases, satisfying the inequality
\begin{align} \label{gammades}
\eta_k \leq \left[ \frac{(1-\frac{1}{d})^k \text{vol}(p)}{\zeta^d \text{vol }B_d(\b0,1)} \right]^{\frac{1}{d}}.
\end{align}
Thus, for each sequence from polytope $p$ to leaf sub-polytope $q_k$ of the binary tree, we can find a sufficiently large $k$ such that $d \eta_k \leq \frac{\gamma_p}{2}$, where $\gamma_p$ is the radius of the smallest enclosing ball of $p$. We note that polytope $q_k$ is enclosed by a ball of radius $d \eta_k$ because $\eta_k$ represents the radius of the primary principle axis of the maximal inscribed ellipsoid of $q_k$. 
The number of hyperplanes required to create the path from $p$ to $q_k$ is equal to the depth of the path, with each hyperplane needing $d+1$ parameters. Consequently, the total number of parameters needed to reduce the enclosing radius from $\gamma_p$ to $\frac{\gamma_p}{2}$ for the path is $k(d+1)$. A binary tree of depth $k$ has $2^k$ nodes, each node is associated with a hyperplane, to reduce the enclosing radius of $p$ to be half of the value for all paths in the tree needs as least $2^k (d+1)$ parameters.

We use the symbol $k_p$ to represent the depth of the binary tree of the polytope $p$, where the smallest enclosing radius of any polytope in the leaf node of the tree is not greater than $\frac{\gamma_p}{2}$. A binary tree of depth $k_p$ contains $2^{k_p}-1$ internal nodes, each associated with a hyperplane. Consequently, reducing the enclosing radius of $p$ to half its original value requires approximately $2^{k_p} (d+1)$ parameters. 
Each partitioning region in $P_l^0$ has its own specific $k_p$ value. Consequently, we can define $\beta_l$ as the maximum value of $k_p$ for $p \in P_l^0$, and $\beta$ as the maximum of $\beta_l$ for $l \in [L]$. Using $\beta$, the number of partitioning polytopes in $P_{l+1}^0$ is $|P_{l+1}^0| \leq 2^\beta |P_l^0| \leq 2^{l\beta}$ for all $l \in [L]$ with $|P_0^0| = 1$. The number of parameters in the construction to generate $P_{l+1}^0$ from $P_{l}^0$ is $(d+1) 2^{l \beta}$. The smallest enclosing radius of $P_0^0$ is $1/2$, and that of $P_l^0$ is $1/2^{l+1}$.

The number of parameters in $\mathcal M_{l+1}^0$, denoted as $\#W_{l+1}$, is equal to the number of parameters in $\mathcal M_l^0$, denoted as $\#W_l$, plus the number of parameters required to obtain $P_{l+1}^0$ from $P_l^0$. This relationship can be expressed as an inequality:
\[ \#W_{l+1} \leq \#W_{l} + (d + 1) 2^{l\beta}. \]
Using the induction $\#W_l \leq (d+1) 2^{l \beta}$, we can derive the relationship $\#W_{l+1} \leq 2 \#W_l \leq (1 + \alpha) \#W_l$, where $\alpha \geq 1$. Substituting the value of $\alpha$ into Eq. (\ref{gammaW}), we obtain $\gamma_s \leq \frac{1}{\#W_L}$ for the number of network parameters of Eq. (\ref{series}) to obtain partition $P_L^0$ with an enclosing radius smaller than $\gamma_s$.

Finally, we remark on constructing the affine linear mapping $M_{l+1}$. We start by obtaining $\mathcal M_l^0$ and then defining the partition induced by $P_l^0$ based on the activation states of the ReLUs in $\mathcal M_l^0$ \cite{hwang2019rectifying}. Each partitioning polytope in $p \in P_l^0$ corresponds to a unique sequence of ReLU activation states, allowing for a direct mapping between $p$ and a sequence of ReLU activations in $\mathcal M_l^0$. This mapping enables us to use the ReLU values in $\mathcal M_l^0$ to identify any partitioning polytope in $P_l^0$. By setting the activation function states corresponding to polytope $p$, $\mathcal M_l^0$ calculates the function restricted to the polytope in the input domain and the sequence of hyperplanes passing anchor points according to the earlier method to refine the polytope. Following this, we can create an array of hyperplanes generated by a sequence of cuts for every partitioning polytope in $P_l^0$. The affine linear mapping $M_{l+1}$ is then obtained as the array resulting from stacking the arrays of hyperplanes generated by cutting each partitioning polytope in $P_l^0$.

\section{Sample size efficiency and benign overfitting} \label{sec:impact}

Our analysis can help clarify the experimental findings related to the learning efficiency of regression and classification functions, as well as the phenomenon of benign overfitting.

\subsection{Sample size efficiency of learning regression and classification tasks}

Comparing the sample sizes required for regression and classification networks to achieve a desired level of consistency represented by $\epsilon$ and $\delta$ allows us to understand the efficiency needed for learning each task. Our analysis, comparing the order of sample complexity given in Corollaries \ref{samplecomplexityR} and \ref{samplecomplexityC}, demonstrates that learning classification to achieve a reliable generalization accuracy $(\epsilon, \delta)$ requires fewer samples than regression. 

Understanding the prior information $K_f$ and $|\partial f|$ falls under PAC-Bayesian analysis, ultimately improving sampling complexity in reaching generalization errors. Our study suggests we can achieve a higher sample complexity by bounding $K_f$ and $|\partial f|$ instead of their precise values.

In Eq. (\ref{regressionm0}), the required sample size for a regression function is determined, indicating that the training data needed for learning a regression function increases exponentially with the input dimension if the Lipschitz constant of $f$ exceeds 1. Conversely, according to Eq. (\ref{classificationm0}), the sample size for the classification function suggests that learning a simple classification function (with a smaller $|\partial f|$) necessitates less training data for the same level of accuracy and confidence.

Inferred from equations (\ref{regressionm0}) and (\ref{classificationm0}), the sample size required for regression is approximately raised to the power of $d-1$ compared to that for classification to achieve equivalent accuracy. Consistent with the findings in \cite{hayou2019mean,huang2020deep,radhakrishnan2023wide}, when employing extra-wide and extra-deep networks to achieve a similar level of consistency, our study indicates that a regression network must contend with substantial training data sizes and effectively manage its Lipschitz function to achieve accuracy and confidence levels comparable to those of a classification network.

\subsection{Benign overfitting}

Research has shown that deep neural networks can achieve remarkably low test errors while interpolating the training data, a phenomenon known as benign overfitting. This discovery challenges the traditional belief that a slight empirical error bias in a learning model will inevitably lead to significant prediction variance, resulting in overfitting. The benign overfitting phenomenon presents an intellectual challenge for understanding the generalization behavior of over-parameterized models. It has also led to the development of fundamental models that utilize modules such as segmentation, classification, and detection for various applications. Training the fundamental models to achieve benign overfitting is crucial for high performance in downstream applications. Additionally, recent studies have demonstrated that the benign overfitting phenomenon is not unique to deep neural networks but can also occur in kernel machines \cite{belkin2018understand}, high-dimensional ridgeless least squares interpolation \cite{hastie2022surprises}, shallow convolutional neural networks \cite{cao2022benign}, and linear regression of over-parameterized models \cite{bartlett2020benign}. 
Our research has practical implications. It reveals that over-parameterization models and densely arranged training data can lead to benign overfitting, a highly desirable outcome. Furthermore, our findings determine the complexity levels needed for regression and classification tasks to achieve benign overfitting.

The equations (\ref{rmse}) and (\ref{gen2-normc}) contain a term on the right-hand side that limits the difference between generalization and empirical errors. This term, known as overfitting, helps determine the level of overfitting a learning algorithm exhibits. Our research indicates that when achieving benign overfitting, the choice of input data density function is not a critical factor. Instead, the availability of ample training data points is crucial, as it links the layout of training points to $\gamma_s$ in the equations. Interestingly, in regression, the overfitting term is proportional to $\gamma_s$, whereas in classification, it is proportional to $\gamma_s^{d-1}$. Connecting $\gamma_s$ to the number of network parameters using Eq. (\ref{gammaW}), it is easier to achieve benign overfitting in classification learning than in regression.

In regression networks, the overfitting term also depends on the Lipschitz constants of the real-valued function $f$ and the network $\mathcal M$. Overfitting is less likely when $f$ has a small Lipschitz constant.
For classification networks, the extent of overfitting also depends on the length of the classification boundary $|\partial f|$. Simple classification functions with shorter boundary lengths are less prone to overfitting. In practice, multi-label classifiers tend to have longer boundaries than binary classifiers, making reducing the generalization error for multi-label classifiers more challenging.

Our analysis relies on the assumption that every covering ball within the input domain needs to encompass at least one training point and an adequate number of training points. The latter assumption is crucial in preventing overfitting in a classification system. This assumption is particularly relevant when considering the observed overfitting in the tree hypothesis class, where each partitioning region of a classification tree consists of precisely one training point.

\section{Discussion} \label{sec:discussion}

\subsection{Shallow versus Deep NNs for learning Lipschitz functions}

Recent research suggests that images of the physical world often have structured compositions and hierarchies. For example, Mhaskar et al. showed in their study \cite{mhaskar2017and} that deep neural networks can effectively represent a target function with hierarchical and compositional structures using fewer parameters than shallow networks. This result can lead to lower generalization errors, as demonstrated in the PAC analysis with finite hypothesis space by \cite{anthony1999neural} (Theorem 16.3), where a finite number of bits encodes each network weight. 
Additionally, comparing the number of regions partitioned by deep neural networks to shallow networks using the same number of activation units can indicate their function approximation capability. Functions that partition the input space into many regions are considered more complex or have better representation power. Research has shown that deep neural networks can partition the input space into exponentially more regions than their shallow counterparts using the same number of ReLU activation units. This point is highlighted in one pioneering work by Pascanu et al. \cite{pascanu2013number} and in the improved bound for deep neural networks by Montufar et al. \cite{montufar2014number} and Serra et al. \cite{serra2018bounding}. According to our analysis, these results imply that the regions generated by deep neural networks can have smaller enclosing balls than their shallow counterparts, ultimately leading to smaller generalization errors.

Here, from a different perspective, we will demonstrate that deep neural networks tend to have a lower Lipschitz constant, resulting in lower generalization errors for regression function, as indicated by Lemma \ref{generalizationbound}.
Using directed acyclic graphs (DAGs) to represent deep neural networks' hierarchical and compositional structures effectively was initially introduced in \cite{mhaskar2016deep}. We refer to these DAGs as DAG-DNNs. Further exploration of this concept was conducted in a study by Hwang et al. \cite{hwang2023analysis}. The study illustrates that DAG-DNNs rely on three primary building modulus operations, series, parallel, and fusion connections, to achieve stability against small input perturbations regarding the network's Lipschitz constant. The nodes in a DAG-DNN perform resizing operations, which enhance the efficiency of function computations by transforming vectors into tensors or vice versa. In DAG-DNNs, the edges are linked to specific computational elements, including affine linear mappings, linear functions, and non-linear activations. The function associated with any pair of connected nodes can be methodically defined by breaking down the DAG-DNN into a series of smaller networks, with one node serving as the input and the other as the output  \cite{hwang2023representation}.

In this analysis, we explore the benefits of going deeper into a network rather than going wider by examining the Lipschitz constants of the three primary building modulus of DAG-DNNs. 
Suppose we have a set of Lipschitz functions, denoted by $k_i$ where $i$ can range from $1$ to $m$. Each function has a Lipschitz constant $l_i$, which is less than or equal to $1$. Combining functions $k_1$ and $k_2$ in a series connection, we obtain the composite function $k_2 \circ k_1$.
The Lipschitz function of module $k_2 \circ k_1$ is
\begin{align*}
\| k_2 \circ k_1 (\bx_1) -  k_2 \circ k_1( \bx_2) \|_2 \leq l_2 \| k_1(\bx_1) - k_2 (\bx_2) \|_2 \leq l_2 l_1 \|\bx_1 - \bx_2\|_2.
\end{align*}
The module parallel connection generates the output vector $\begin{bmatrix} k_1 (\bx) \\ \vdots \\ k_m(\bx) \end{bmatrix}$ from the input vector $\bx$. The Lipschitz constant of the parallel module is given by the expression $\sqrt m \max_i \{ l_i\}$, as shown in the equation 
\begin{align*} 
\left\|\begin{bmatrix} k_1 (\bx_1)-k_1(\bx_2)  \\ \vdots \\ k_m(\bx_1) - k_m(\bx_2) \end{bmatrix}\right\|_2^2 = \sum_{i=1}^m  \| k_i (\bx_1)-k_i(\bx_2)\|^2_2 \leq \sum_{i=1}^m l_i^2 \| \bx_1 - \bx_2\|_2^2 \leq m \max_i \{l_i^2\} \| \bx_1 - \bx_2\|_2^2.
\end{align*}
The fusion module connection involves merging the inputs of $m$ channels with $\{\bx_1,  \cdots,  \bx_m \}\rightarrow \begin{bmatrix} k_1 (\bx_1) \\ \vdots \\ k_m(\bx_m) \end{bmatrix}$. The Lipschitz constant of the fusion module is determined by the maximum value among $l_i$ values. This is because:
\begin{align*} 
\left\|\begin{bmatrix} k_1 (\bx_1)-k_1(\bx_1')  \\ \vdots \\ k_m(\bx_m) - k_m(\bx_m') \end{bmatrix}\right\|_2^2 = \sum_{i=1}^m  \| k_i (\bx_i)-k_i(\bx_i')\|^2_2 \leq \sum_{i=1}^m l_i^2 \| \bx_i - \bx_i'\|_2^2 \leq \max_i \{l_i^2\} \sum_{i=1}^m \| \bx_i - \bx_i'\|_2^2.
\end{align*}

Shallow networks have only one layer of interconnected modules, while deep networks have multiple layers appended to one another. When comparing the Lipschitz constants of the building modules of DAG-DNNs, it is noted that deep neural networks composed of a sequence of series connection modules can generate functions with smaller Lipschitz constants from training data. This understanding is essential in comprehending the advantage of deep neural networks over shallow ones in developing networks with smaller Lipschitz constants, a significant aspect of generalization error for regression functions.

\subsection{The expected number of balls covering for the first time with a continuous density function} \label{expectednumber}

When dealing with a continuous function $\mathcal P$, we can establish abound for the minimum expected value of $\mathcal E(X_{min})$ by leveraging the fact that any continuous function can be approximated to arbitrary accuracy by a piece-wise constant function over a lattice grid structure of sufficiently small size, tailored to the specific function and accuracy requirements. Once we determine $\gamma_s$ for a specific accuracy to approximate the function $\bar l$, which is continuous because of $\mathcal P$ being a continuous function, we can construct this approximation using a Haar basis with a support of $d$-cube.

Let $\text{cube}_p$ represent the cube that is maximally inscribed in ball $p$, and let $k_p = \max_{\bx \in \text{cube}_p} \bar l(\bx)$ be the constant value that approximates $\bar l(\bx)$ for $\bx$ in $\text{cube}_p$. We can obtain the approximation:
\begin{align}
\int_{\text{cube}_p} \bar l(\bx) \mathcal P(\bx) d\bx = \int_{\text{cube}_p}  \frac{1}{P(\bx; B_d(\bx, \gamma_s))} \mathcal P(\bx) d\bx \leq k_p \int_{\text{cube}_p} \mathcal P(\bx) d\bx \leq l,
\end{align}
where $l$ is a finite constant uniformly for all $\text{cube}_p$ due to the continuous function on a compact domain, and the number of non-overlapping $d$-cubes covering the unit $d$-ball is finite. The volume of the largest $d$-cube that can fit inside a ball of radius $\gamma_s$ in $\mathbb{R}^d$ is given by the formula $\frac{(2\gamma_s)^d}{d^{d/2}}$. Therefore, the length of each side of the cube is $\frac{2 \gamma_s}{\sqrt{d}}$, and the minimum number of such cubes needed to cover the unit $d$-ball is $\frac{1}{\gamma_s^d}(\frac{\sqrt{d}}{2})^d$. The $\mathcal E(X_{min})$ value is approximately equal to the number of cubes in the approximation of function $\bar l$. Thus, we have:
\begin{align}
\mathcal E(X_{min})  = \int_{\bx \in B_d(\b0,1)} \bar l(\bx) \mathcal P(\bx) d\bx = \sum_{\text{cube}_p} \int_{\text{cube}_p} \bar l(\bx) \mathcal P(\bx) d \bx \leq \frac{l}{\gamma_s^d}(\frac{\sqrt{d}}{2})^d.
\end{align}

\subsection{Inductive bias related to our analysis}

When dealing with over-parameterized systems, the inductive bias serves as a constraint that limits the potential solutions of a learning algorithm. Our generalization bounds for regression and classification can be attributed to different prior knowledge.

In the context of regression, when we lack knowledge of the exact Lipschitz constant of the target function, we can impose constraints on the norm of the weight coefficients, which is related to the Lipschitz constant of the network. Our analysis indicates that the generalization bound favors a network of low Lipschitz constant. Additionally, we acknowledge the potential regularization brought about by the gradient method due to the connection discussed in Section  \ref{sec:review} between the $l_2$-regularized solution, which is connected to the Lipschitz constant, and the gradient method.

When considering the classifier, our analysis suggests that the appropriate focus should not be solely on the magnitude of the weight coefficients but instead on the classification boundary of the target classifier $f$, denoted as $\partial f$. A smooth classification boundary is preferred as it helps in minimizing generalization errors. To estimate the unknown $\partial f$, we can use the classification boundary of $\mathcal M$ and constrain it using norms related to the function derivatives of $\mathcal M$, denoted as $\| \partial M\|$. When an optimization method is used to reduce the empirical classification error, the classification boundary of $\mathcal M$ becomes close to that of $f$, resulting in $\| \partial M \| \approx \| \partial f\|$.

\subsection{Efficient algorithms relating covering geometry and network architecture}

In Section \ref{oracle}, we present a deep neural network that links the radius of covering balls, denoted as $\gamma_s$, induced by the network, to the network's parameter count $\#W_L$, with  $\gamma_s$ being propositional to $\frac{1}{\#W_L}$. While it is possible to achieve this in polynomial time, the construction is not programmatically efficient and is contingent on the geometry of polytopes in the input domain's partition. What kind of hyperplane arrangement in network layers would enable the efficient implementation of the above relation? Additionally, what network structure can result in a decay rate faster than $\frac{1}{\#W_L}$? These questions merit further investigation as they could potentially lead to developing an efficient new neural network architecture with a few parameters and a training algorithm with a harnessed generalization error.

\subsection{Sample complexity for generative models}

The sample complexity evaluation determines the number of sampling points needed to approximate an unknown density distribution function. Our approach uses the 2-norm measurement for regression estimation, described by a specified geometric-related parameter ($\gamma_s$) on sampling points from the unknown density.

In a recent research paper \cite{kadkhodaie2023generalization,song2020score}, the authors explored using a diffusion generative model \cite{sohl2015deep} for density estimation. This method involves applying a noise diffusion process to measure density variances using the KL divergence. Following this perspective, well-known generative models using Jensen-Shannon divergence \cite{goodfellow2020generative}, Wasserstein-GAN \cite{arjovsky2017wasserstein}, and variational auto-encoder \cite{kingma2013auto} can be seen as density estimation techniques that use different algorithms to approximate the target density with density generated using sample points. Inspired by this viewpoint, we pose the question of further study on whether our geometrically adaptive construction approach could serve as the basis for a generative model and whether it is feasible to compare generative models using quantitative measures of sample complexity.

\bibliographystyle{ieeetr}
\bibliography{DeepRef}

\end{document}